\documentclass[lettersize, journal]{IEEEtran}
\usepackage{amsmath,amsfonts}
\usepackage{algorithmic}
\usepackage{algorithm}
\usepackage{array}
\usepackage[caption=false, font=normalsize, labelfont=sf, textfont=sf]{subfig}
\usepackage{textcomp}
\usepackage{stfloats}
\usepackage{url}
\usepackage{verbatim}
\usepackage{graphicx}
\usepackage{caption}
\usepackage{float}
\usepackage[square, sort, comma, numbers]{natbib}
\usepackage{makecell}
\usepackage{soul}

\title{A Sentinel-2 multi-year, multi-country benchmark dataset for crop classification and segmentation with deep learning}
\author{Dimitrios Sykas, Maria Sdraka, Dimitrios Zografakis, Ioannis Papoutsis \\
        \normalsize Institute of Astronomy, Astrophysics, Space Applications \& Remote Sensing, National Observatory of Athens \\
        \normalsize \{dimsyk, masdra, dimzog, ipapoutsis\}@noa.gr
        }
             
\begin{document}
    
    \maketitle
    
    \begin{abstract}

In this work we introduce Sen4AgriNet, a Sentinel-2 based time series multi country benchmark dataset, tailored for agricultural monitoring applications with Machine and Deep Learning. Sen4AgriNet dataset is annotated from farmer declarations collected via the Land Parcel Identification System (LPIS) for harmonizing country wide labels. These declarations have only recently been made available as open data, allowing for the first time the labeling of satellite imagery from ground truth data. We proceed to propose and standardise a new crop type taxonomy across Europe that address Common Agriculture Policy (CAP) needs, based on the Food and Agriculture Organization (FAO) Indicative Crop Classification scheme.
Sen4AgriNet is the only multi-country, multi-year dataset that includes all spectral information. It is constructed to cover the period 2016-2020 for Catalonia and France, while it can be extended to include additional countries. Currently, it contains 42.5 million parcels, which makes it significantly larger than other available archives. We extract two sub-datasets to highlight its value for diverse Deep Learning applications; the Object Aggregated Dataset (OAD) and the Patches Assembled Dataset (PAD). OAD capitalizes zonal statistics of each parcel, thus creating a powerful label-to-features instance for classification algorithms. On the other hand, PAD structure generalizes the classification problem to parcel extraction and semantic segmentation and labeling. The PAD and OAD are examined under three different scenarios to showcase and model the effects of spatial and temporal variability across different years and different countries. The dataset can be accessed in: \url{https://sen4agrinet.space.noa.gr}
\end{abstract}


    \section{Introduction}
Over the past years, Copernicus Sentinels (1 \& 2) and NASA’s Landsat satellites have been consistently collecting images harmonized in the spectral, temporal and spatial dimensions. The free and open distribution of such an imagery archive has enabled, among others, the consistent and robust monitoring of agricultural activities. Furthermore, the recent developments in Artificial Intelligence (AI) algorithms and models has propelled the adoption and implementation of novel Machine Learning techniques, paving the way for a more efficient modeling of the complex agricultural ecosystems and the generation of expertise to support smart farming, Common Agriculture Policy (CAP) implementation, and agricultural insurance. 

One of the major problems faced by researchers in this field is the absence of country-wide labeled data that are harmonized along space and time. Specifically in the EU, the Common Agriculture Policy (CAP) has placed a stepping stone to overcome this issue by legally establishing Paying Agencies in each EU country which are responsible for distributing subsidies to farmers. In order to fulfill their objectives, Paying Agencies systematically collect the cultivated crop type and parcel geometries for every farmer and record it via the Land Parcel Identification System (LPIS) \cite{eca-land-id-system} in a standardized way for each country. Unfortunately, public access to these farmer declarations has been restricted for several years, thus making it almost impossible to get country-wide ground truth data. However, since 2019 and for the first time these datasets are gradually becoming open (e.g. France, Catalonia, Estonia, Croatia, Slovenia, Slovakia and Luxemburg). This change offers a significant opportunity for the Earth Observation (EO) community to explore novel and innovative data-driven agricultural applications, by exploiting this abundance of new LPIS information.

In principle, this fusion of the LPIS data sources has tremendous potential but there are still some barriers to overcome. First of all, the LPIS system of each country is customly configured to utilize the local language of the crop types and the specific taxonomy structure of the crops that matches the local subsidies policy implementation. This non-standardization of the labels prohibits the spatial generalization of Deep Learning (DL) models and thus needs to be carefully handled to achieve a common representation consistent among countries. On top of these contextual/semantic barriers, parcels are mapped in the corresponding national cartographic projection which in all cases is different from the cartographic projection of the satellite images and pose an additional challenge on the preparation of a consistent, proper and at scale DL-ready dataset.

\begin{figure*}[htb]
    \centering
    \includegraphics[width=0.80 \textwidth]{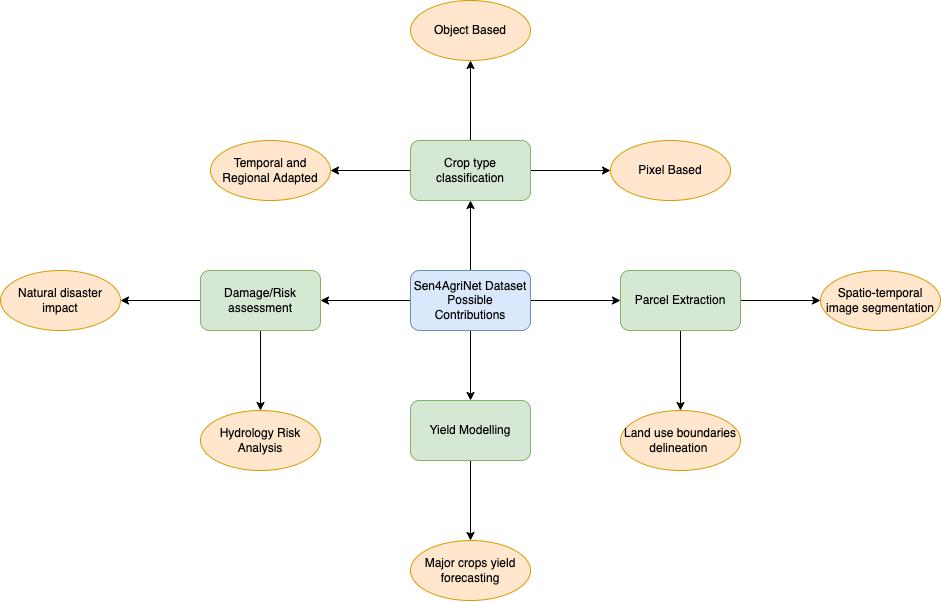}
    \caption{Taxonomy of Sen4AgriNet potential applications.}
    \label{fig:potential_apps}
\end{figure*}

In this work we introduce, present and use Sen4AgriNet, a unique benchmark EO dataset for agricultural monitoring with the following key characteristics: a) it is pixel based to capture spatial parcel variability, b) it is multi-temporal to capture the crop phenology phases, c) it is multi-annual to model the seasonal variability, d) it is multi-country to model the geographic spatial variability, e) it is object-aggregated to further incorporate ground truth data (parcel geometries) in the process, and f) it is modular since it can be enlarged with parcels from more EU countries or expanded in a straightforward way to include additional sensor and non-EO data (e.g. meteorological data). A preliminary version of Sen4AgriNet was  introduced and discussed in~\cite{9553603}. This paper provides an extended version for Sen4AgriNet, provides additional technical details on its construction, and includes a series of new experiments on top of the dataset. In Fig. \ref{fig:potential_apps} we provide a diagram presenting potential applications for Sen4AgriNet, highlighting the major challenges for which the dataset can be used to address. Starting from crop type classification and parcel extraction with their variants, the dataset can also be used for yield modelling applications, where strategic plans for efficiency and sustainability can be investigated, as well as damage/risk assessment tasks for successful planning of relief and regeneration actions.

The Sen4AgriNet contains approximately 225,000 5-year multitemporal Sentinel-2 patches co-registered with open LPIS data for regions in Spain and France with a total size of 10TB. The dataset is splitted into two distinct sub-datasets: the Patch Aggregated Dataset (PAD) and the Object Aggregated Dataset (OAD). PAD contains the original patches of Sentinel-2, i.e. the raster reflectance bands as time series. OAD is built on top of PAD, by aggregating raster values at parcel level, thus producing one sample per parcel which contains all the averaged spectral values with statistics for all bands and available time stamps.

The motivation for creating this dataset is based on the existing issues and challenges that the agriculture remote sensing domain has faced over the last years, despite the access to open satellite data archives. Namely, we aim to provide an outlet for standardized label nomenclature, connection of crop types to operational usage of the results (e.g. EU CAP) and enhancement of the generalization ability of the trained classifiers.
Our main contributions can be summarised as follows:
\begin{itemize}
    \item We develop and deliver Sen4AgriNet, a benchmark dataset tailored for Machine Learning (ML)/Deep Learning (DL) that can be used for a variety of EO based applications, such as crop type classification, parcel extraction, parcel counting and semantic segmentation.
    \item We introduce a unified crop taxonomy based on the Food and Agriculture Organization (FAO) Indicative Crop Classification scheme, to directly address CAP needs.
    \item We construct and deliver two reduced versions of Sen4AgriNet, one tailored for pixel-based and one for object-based applications. 
    \item We conduct a series of baseline experiments to assess for the first time the generalization capabilities of state-of-the-art DL methods across space (different countries) and time (different cultivation years). 
    \item We open up for reuse Sen4AgriNet, our trained models and our code for generating Deep Learning analysis ready datasets. 
\end{itemize}

This paper is structured as follows: initially we review similar existing datasets in EO designed for ML/DL applications. Then the methodology for developing Sen4AgriNet is presented, including the proposed taxonomy and its main classes structure. Third, we present the experimental designs and DL model architectures used to test different generalisation scenarios in the context of the pixel and object based datasets. Finally, the results of the experiments are presented and their implications are briefly discussed.

    \section{Remote Sensing Datasets for ML Applications}

The concept of creating reference datasets in remote sensing problems targeted at AI algorithms to solve scientific questions has recently gained traction in the community. Most of the existing datasets focus on problems related to land use/land cover classification exploiting pre-existing open data sources. One of the first attempts in this domain is BigEarthNet \cite{Sumbul_2019}, a large-scale benchmark archive of 125 Sentinel-2 tiles corresponding to acquisitions over 10 European countries from June 2017 to May 2018. These tiles were atmospherically corrected with sen2cor \cite{sen2core} and splitted into 590,326 non-overlapping image patches to better address computer vision problems. For each created patch the multiple corresponding land cover classes were subsequently exported via the CORINE Land Cover (CLC) database~\cite{CLC2018} and added as label annotations, establishing BigEarthNet a good fit for multi-class, multi-label image classification applications.

A similar approach to BigEarthNet is presented in \cite{helber2019eurosat} (Eurosat), where a multi-class annotation dataset with all 13 spectral bands of Sentinel-2 is proposed. Eurosat consists of 10 land cover classes with 27,000 labeled and geo-referenced image patches. A more specialized dataset named so2Sat is presented in \cite{9014553}. It focuses on urban areas and classes across the planet combining both Sentinel-1 and Sentinel-2 image patches. Key asset of the dataset is the manual labeling process which was designed and perfomed by 15 domain experts.

More relevant from a domain scope to our proposed dataset is the ``CV4A Kenya Crop Type Competition'' dataset~\cite{radiant-earth-foundation}. It combines the temporal aspect of satellite image acquisition with crop types. The dataset uses both the multi-temporal coverage of satellite images and the spatial distribution of farm holdings. Key shortcoming of this dataset are the small number of agricultural parcels, the small number of different crop type classes, and the fact that the included satellite and label data cover a single year and no multi-annual data are recorded. 

Another relevant to the agricultural domain dataset is reported in \cite{breizhcrops2020}. The BreizhCrop dataset is used for the supervised classification of field crops from satellite time series and consists of combined labeled data and Sentinel-2 top-of-atmosphere as well as bottom-of-atmosphere time series in the region of Brittany, north-east France, spanning throughout 2017. Key characteristic of the dataset is the object based approach, i.e. each parcel is represented as one observation (spatial aggregation) with several features (Sentinel-2 reflectances over different timesteps).

In \cite{tong2019landcover} a large-scale land-cover dataset with Gaofen-2 (GF-2) satellite images was created. The Gaofen Image Dataset (GID) focuses on land cover classes, with special remarks on the coverage size and spatial distribution resolution. GID consists of two parts: a large-scale classification set and a fine land-cover classification set. The large-scale classification set contains 150 pixel-level annotated GF-2 images, and the fine classification set is composed of 30,000 multi-scale image patches coupled with 10 pixel-level annotated GF-2 images. The training and validation data with 15 categories is collected and re-labeled based on the training and validation images with 5 categories respectively.

A recently published dataset that resembles the one presented in this study is PASTIS \cite{garnot_panoptic_2021}, which includes time series of multispectral images obtained from the Sentinel-2 satellite constellation. It contains the 10 non-atmospheric bands resampled to the highest spatial resolution of 10m and the labels expand over 18 crop types. However, contrary to Sen4AgriNet, PASTIS provides single-year observations in a single country, thus limiting the variance and diversity of the data.

From the reported datasets, most of them do not take into account the time series of satellite image acquisition, while the majority also focuses on annotating tags instead of masks, which constrains the usage of the dataset in simple classification scenarios (Table \ref{table:existing-datasets}). For example image segmentation, object detection, parcel counting, etc. are applications that cannot be tackled with this kind of aggregation. In addition, removing the temporal aspect of satellite time series restricts Deep Learning models from learning the temporal/seasonal dynamics of the classes. Of course, the temporal dynamics in classes are not always applicable, e.g. in land cover types and urban structures. On the contrary, crop type classes show significant spectral and spatial changes over time, thus it is essential to include the time aspect for such applications.

\begin{table}[htb]
    \caption{List of state-of-the-art remote sensing datasets for ML applications. Sen4AgriNet is the only dataset that cumulatively supports both pixel and object based aggregations, spans across different countries and multiple years, its input satellite data is time-series, and contains several millions of annotations. The proposed Sen4AgriNet dataset is marked in bold.}
    \label{table:existing-datasets}
    \centering
    \begin{tabular}{ >{\raggedright\arraybackslash} m{14mm} m{8mm} m{14mm} m{8mm} m{7mm} m{7mm}}
    \hline
    \multicolumn{1}{>{\raggedright\arraybackslash}m{14mm}}{\textbf{Dataset}} & \multicolumn{1}{>{\raggedright\arraybackslash}m{8mm}}{\textbf{Type}} & \multicolumn{1}{>{\raggedright\arraybackslash}m{14mm}}{\textbf{Area}} &
    \multicolumn{1}{>{\raggedright\arraybackslash}m{8mm}}{\textbf{Span}} &
    \multicolumn{1}{>{\raggedright\arraybackslash}m{7mm}}{\textbf{Use}} & \multicolumn{1}{>{\raggedright\arraybackslash}m{7mm}}{\textbf{Labels}} \\
    \hline
        BigEarthNet & Pixel & EU & None & LC & 590k \\
        Eurosat & Pixel & EU & None & LC & 27k \\
        so2Sat & Pixel & Global & None & LC & 500k \\
        CV4A & Object & Kenya & 1 Year & Urban & 4k \\
        BreizhCrop & Object & France & 1 Year & Agri. & 610k \\
        GID & Pixel & China & N/A & LC & 30k \\
        PASTIS & Both & France & 1 year & Agri. & 124k \\
        \textbf{Sen4AgriNet} & \textbf{Both} & \textbf{Cat \& Fr} & \textbf{Annual} & \textbf{Agri.} & \textbf{42m} \\
    \end{tabular}
\end{table}

Adding to the complexity of the problem, crop types have variable spectral and spatial response at different geographic areas and cultivation techniques, which are usually strongly connected to the geographic regions. Phenology stages, different seeding dates and agricultural practices, and varying meteorological conditions are just a few parameters that affect the spectral signatures of the satellite data, even for the same crop type. Therefore constructing efficient, accurate and robust DL models requires a domain adaptation strategy. The Sen4AgriNet dataset was designed in order to foster the development of such machine learning approaches.

    \section{Crop Type Classification with Deep Learning}

Several DL approaches have been recently proposed for the classification of crop types on optical satellite imagery, outperforming established methods based on computer vision or traditional Machine Learning techniques, e.g.~\cite{sitokonstantinou2018scalable}. For example, in \cite{7891032} an ensemble of 1D and 2D Convolutional Neural Networks (CNN) is employed in order to discern 11 types of land cover from Sentinel-2 and Landsat-8 images over Ukraine. The predictions are further refined by fusing auxiliary information such as parcel boundaries, statistical data, vector geospatial data, and more. In \cite{rs11171986} a more sophisticated architecture is proposed (\emph{FG-Unet}) which is based on the popular U-Net model \cite{unet}, extended with a second branch for independent classification of every image pixel enabling the model to produce coarser polygon boundaries. The input of this method is a window of three Sentinel-2 images over France captured at three different time steps.

A number of publications have also explored 3D convolutions to simultaneously handle the spatial, spectral and temporal components of the input images. For example, in \cite{rs10010075} the convolutional layers of a VGG model are replaced with their 3D equivalents in order to classify GaoFen-1/2 image pixels into 9 land cover classes. Similarly, in \cite{ZHONG2019111411} MODIS imagery along with NDVI (normalized difference vegetation index) and EVI (enhanced vegetation index) indices are fed to a simple CNN architecture with 3D convolutions for the detection of winter wheat.

On a different note, several studies have regarded the problem of crop type/land cover classification as a time-series classification problem and have proposed architectures specifically designed to exploit the temporal aspect of the input data. Such models perform either in an N-to-N scheme producing a prediction for every input time step, or in an N-to-1 scheme producing a single prediction after examining the whole input sequence. Here we will only consider the latter case as it is similar to the approach followed in this work. Previous studies have incorporated recurrent neural networks (RNNs) in their pipeline, such as the Long Short-Term Memory (LSTM) model \cite{lstm}, which are fed pixel vectors of multiple time steps \cite{ZHONG2019430, rs11141665, XU2020111946, 9026754}. In \cite{INTERDONATO201991} a 2-branch model is proposed (DuPLO) whose first branch is a CNN spatial feature extractor, whereas the second branch is a Gated Recurrent Unit (GRU) \cite{gru} temporal feature extractor. The outputs of both branches are then fused and fed to a final classification layer which makes a prediction for the central pixel of the input window.

Another popular technique for handling spatiotemporal data is the TempCNN model, a custom deep neural architecture which performs convolution simultaneously on the spectral and temporal dimensions, thus managing to extract useful information for the phenological stages and the spectral signature of the different land cover types. This model was first introduced in \cite{rs11050523} for Formosat imagery classification and then subsequently used in \cite{RUWURM2020421} in comparison with LSTM, DuPLO, Multi-scale ResNet \cite{7780459} and Transformer \cite{NIPS2017_3f5ee243} for Sentinel-2 input data. The Transformer was also explored by \cite{9157055} in a pipeline which includes three Multilayer Perceptrons (MLPs) for feature extraction and final classification.

Lastly, a number of studies have also employed convolutional recurrent layers which are essentially RNNs performing 2D convolutional operations internally. An example of such method is shown in \cite{ijgi7040129} where a Convolutional Gated Recurrent Unit (ConvGRU) takes an input sequence of Sentinel-2 images both in correct and reverse order and produces a classification map for 17 crop classes. A more recent module was proposed in \cite{9373965}, named Stackable Recurrent Cell (STAR), and then extended with convolutional layers in \cite{TURKOGLU2021112603} (MS-ConvSTAR) for hierarchical crop type classification on Sentinel-2 imagery. This model achieves faster and more stable convergence than the LSTM/GRU equivalents.

\section{Domain Adaptation}

When there is a significant divergence between the train and test data distributions, a model trained on the former will likely fail to generalize on the latter and thus performance will be greatly degraded. A popular approach which attempts to tackle this problem is Domain Adaptation (DA), a special branch of transfer learning which aims to alleviate the variation between the source (train set) and target (test set) data distributions caused by \emph{data shift}, \emph{concept drift} and \emph{multi-modal domain shift} often observed in Remote Sensing data \cite{campsvalls_deep_2021}. Data shift describes the spectral differences between images captured under different conditions, i.e. different atmospheric effects, sun positions, sensor angles, etc. Concept drift refers to the variation of the intrinsic class characteristics over time and/or space. For example, the same crop displays different spectral signatures as the plant develops, since leaf characteristics and biomass-to-soil ratios can vary both over time (seasonal, phenotypical changes) and space (regional agricultural policies). Finally, multi-modal domain shift arises when multiple sensor types are utilised and differences in bands, resolution, etc are observed. In the present study, multi-modal domain shift is not an issue to be handled since all images were acquired by the Sentinel-2 satellite, but the other two challenges of data drift and concept shift are relevant. In particular, data drift can possibly be observed throughout the whole dataset since atmospheric and illumination conditions are unstable and may vary across time, whereas concept shift is prevalent both in a single examined time series and across regions.

Through Domain Adaptation techniques a model trained on the source domain can be carefully transferred to the target domain without suffering from the effects of the aforementioned distribution variations. There are three major types of DA depending on the availability of ground truth labels in the target domain: (i) Supervised DA, (ii) Unsupervised DA and (iii) Semi-Supervised DA \cite{wang_deep_2018}. In Supervised DA, target data are fully labeled, but labels may be fewer and/or different than those of the source data. In Unsupervised DA, no ground truth labels are available for the target data, whereas in Semi-Supervised DA few labelled instances may be present in the otherwise unlabeled target data set. Several DA approaches have been proposed over the years and can be categorized into three families. Discrepancy-based DA methods attempt to fine-tune the model by minimizing some criterion between the source and target distribution. This criterion can be based on the target labels (e.g. \cite{motiian_unified_2017}, \cite{Gebru_2017_ICCV}, \cite{tong_land-cover_2020}), the statistical shift between the distributions (e.g. \cite{Yan_2017_CVPR}, \cite{sun_deepcoral}, \cite{Damodaran_2018_ECCV}), the architecture of the model (e.g. \cite{Huang_2017_ICCV}, \cite{8310033}, \cite{Xiao_2016_CVPR}) or the geometrical properties of the distributions (e.g. \cite{chopra2013dlid}). Adversarial-based DA methods employ a domain discriminator to assess whether the generated features correspond to the source or the target domain, thus encouraging domain confusion and the production of more robust features (e.g. \cite{Tzeng_2017_CVPR}, \cite{Bousmalis_2017_CVPR}, \cite{NEURIPS2018_ab88b157}). Finally, reconstruction-based DA aims to achieve feature invariance either by translating the target to the source domain (e.g. \cite{Zhu_2017_ICCV}, \cite{tasar_colormapgan}) or by mapping both source and target domains to a common latent space (e.g. \cite{Huang_2018_ECCV}, \cite{lee_drit_2020}).

Specifically for the task of crop type classification, a number of methods have been proposed for cross-region adaptation, such as \cite{wang_phenology_2021} and \cite{martini_domain-adversarial_2021} which take as input a time series of images and output a single segmentation map. A method published recently in \cite{nyborg_timematch_2021} additionally accounts for the phenology shift observed for a single crop between different geographical regions and with a time shift estimation procedure and a semi-supervised learning scheme it manages to boost the performance of the model proposed in \cite{9157055}. \\
    \section{Methodology}

\subsection{Harmonized crop type taxonomy}

The Indicative Crop Classification (ICC)~\cite{alma993810353402676} scheme was developed by the United Nations FAO organization. It is an approach to produce a harmonized vocabulary and taxonomy for crops and plants that are used in food production. The Common Agriculture Policy (CAP) in Europe is an example that requires such normalization among the crop labels, since each country member uses different naming systems and languages. Therefore, Sen4AgriNet adopts and customises an extended version of FAO ICC in order to create a universally applicable crop label nomenclature. 

\begin{figure*}[htb]
    \centering
    \includegraphics[width=0.85 \textwidth]{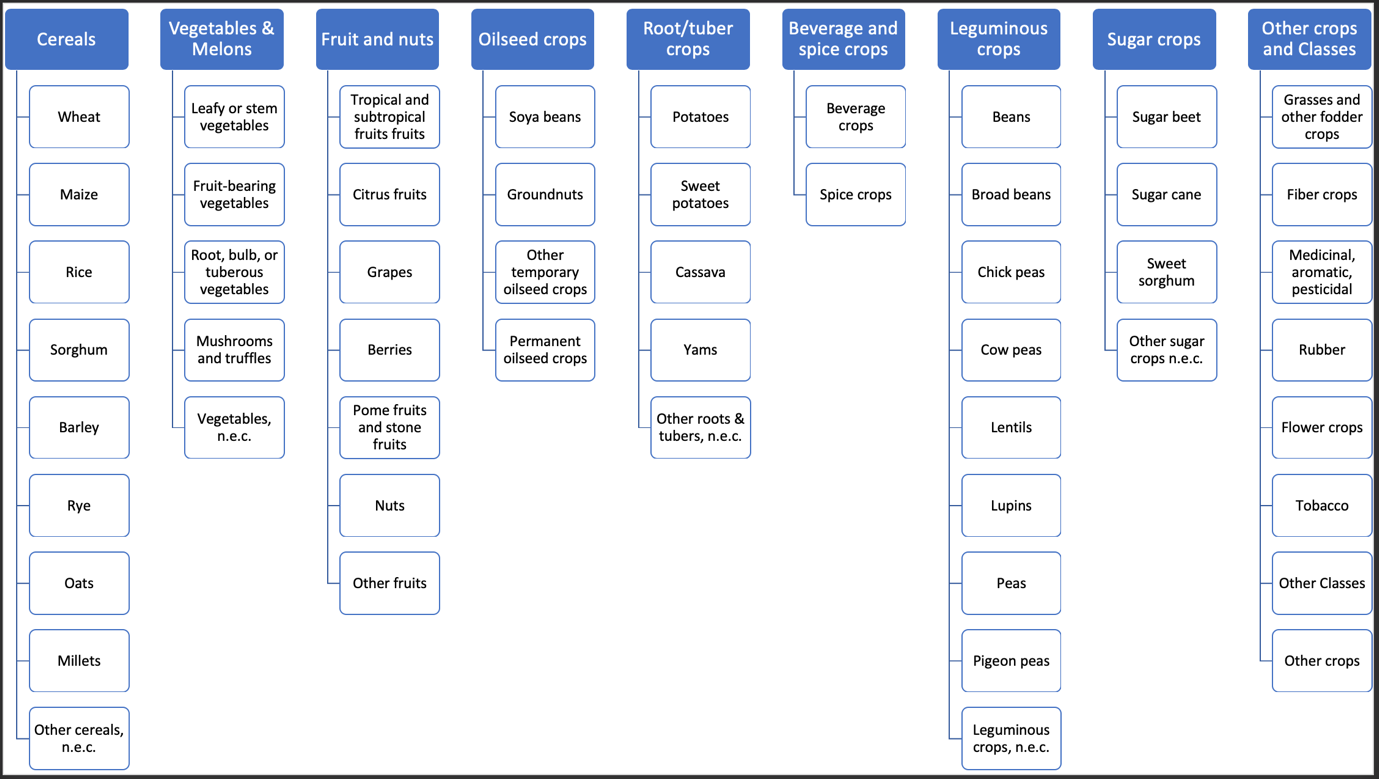}
    \caption{The proposed Sen4AgriNet Crop Categorization taxonomy structure, inspired by FAO and based on the customisation of the ICC~\cite{alma993810353402676} model.}
    \label{fig:fao}
\end{figure*}

A harmonized crop type taxonomy which is designed to be used for satellite crop type classification should not contain solely crop classes, since agricultural parcels co-exist with other unrelated classes in satellite images. Therefore, additional classes are needed to annotate pixels (or objects) that are present in the satellite imagery, but are out of scope from the agricultural domain context. In addition, in order to promote the adoption of the dataset under the CAP regulatory framework, relevant land use classes need to be added. To this end, we created this version of Sen4AgriNet by including some major land use classes from CLC~\cite{CLC2018}, and a few additional classes to complete the semantic concepts (``fallow land'' sub-class, ``barren land'' sub-class and ``no data available'' sub-class). CLC classes related to agriculture and forestry were not included, since they are properly and in detail covered by FAO ICC. 

FAO ICC~\cite{alma993810353402676} has 4 hierarchies (``Group'', ``Class'', ``Sub-Class'', ``Order'') with corresponding numeric ordering. From the 160+ entities in the hierarchy, the fourth one (``Order'') contained only 15 entities. In order to reduce the complexity of any future classification attempts and different label groupings, the ``Order'' level was removed by upgrading the corresponding entities to ``Sub-Class'' level and removing their parent ``Sub-Class''. Adopting and customising this FAO/CLC crop classification scheme to create Sen4AgriNet provides the following benefits:

\begin{enumerate}
    \item{Single language (English) is used and naming for all classes across all participating countries.}
    \item{Classes are normalized among different datasets.}
    \item{Hierarchical class structure is adopted. Depending on the application different levels of classes can be used.}
    \item{Additional non-agricultural classes are used to model RS spectral signatures to support agricultural applications.}
\end{enumerate}

The presented custom FAO/CLC classification scheme has a total of 9 groups, 168 classes and sub-classes. The 161 classes/sub-classes are crop related, 4 are some major CLC classes (as sub-classes in this hierarchy), 2 are the fallow and barren lands, and 1 is the no data sub-class. Please refer to Fig. \ref{fig:fao} for a visualization of the proposed taxonomy.

\subsection{Crop Type Labels - LPIS}

The LPIS datasets that are used to create the Sen4AgriNet dataset undergo a tailored process before using them to annotate Sentinel-2 imagery. Initially, the crop type classes are translated from the local language (Spanish/Catalan and French) into English and mapped to the semantically normalized FAO ICC crop classification scheme. This is a laborious and manual procedure, but essential in order to harmonize the different labels among different LPIS systems. The current version of Sen4AgriNet contains: 

\begin{itemize}
    \item{The LPIS data for the region of Catalonia provided by the ``Agricultura, Ramaderia, Pesca i Alimentació'' with an Open Data Commons Attribution License from 2016 – 2020 \cite{crop-map-catalonia}, with a total of 2.5M parcels.}
    \item{France LPIS data provided by the French Paying Agency with an Open Data Commons Attribution License from 2016 - 2019 \cite{LEVAVASSEUR2016541}, with a total of 40M parcels.}
\end{itemize}

\begin{figure*}[ht]
    \centering
    \subfloat[]{\includegraphics[width=0.20 \textwidth]{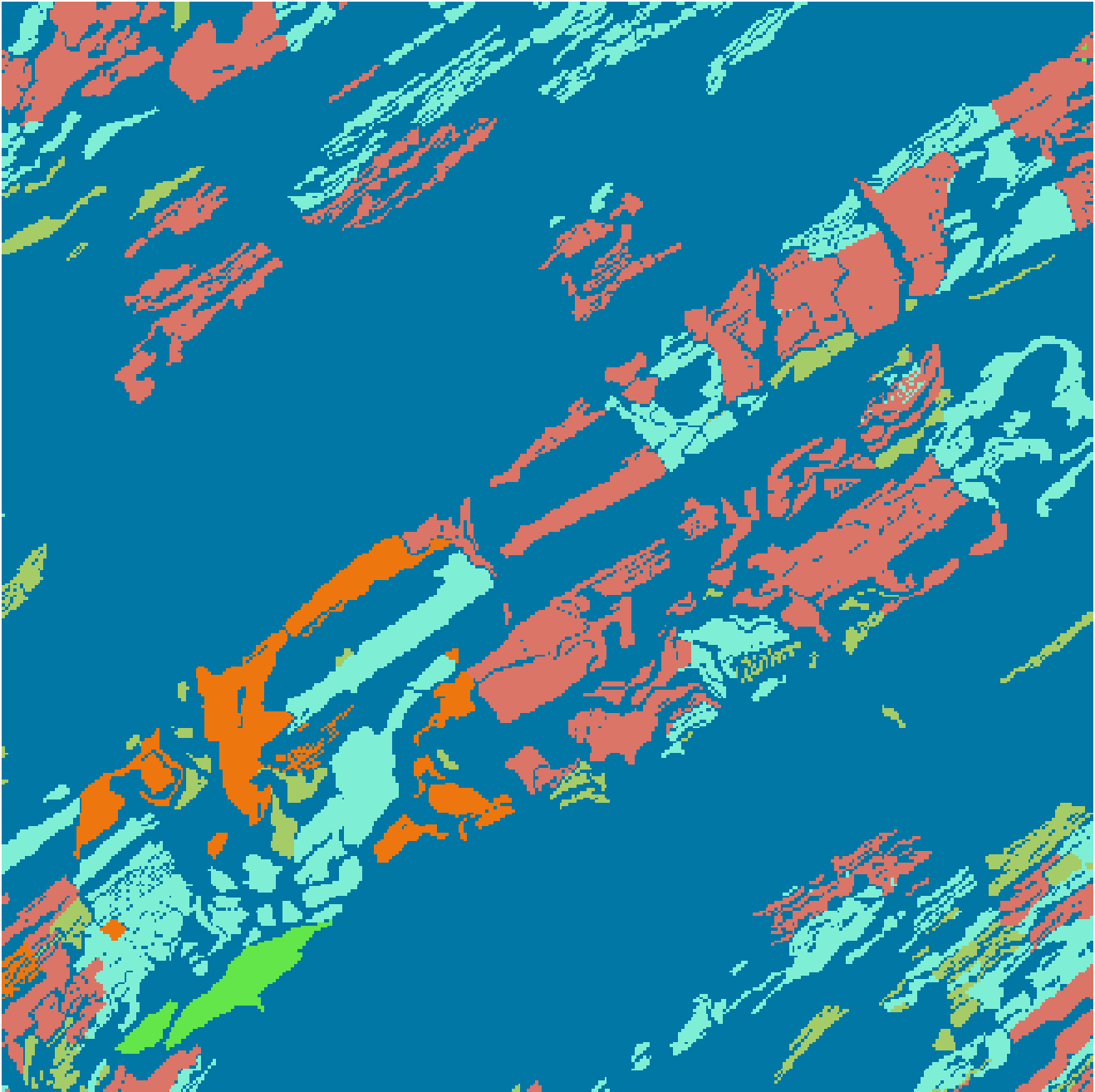}%
    \label{dataset:cat1}}
    \hfil
    \subfloat[]{\includegraphics[width=0.20 \textwidth]{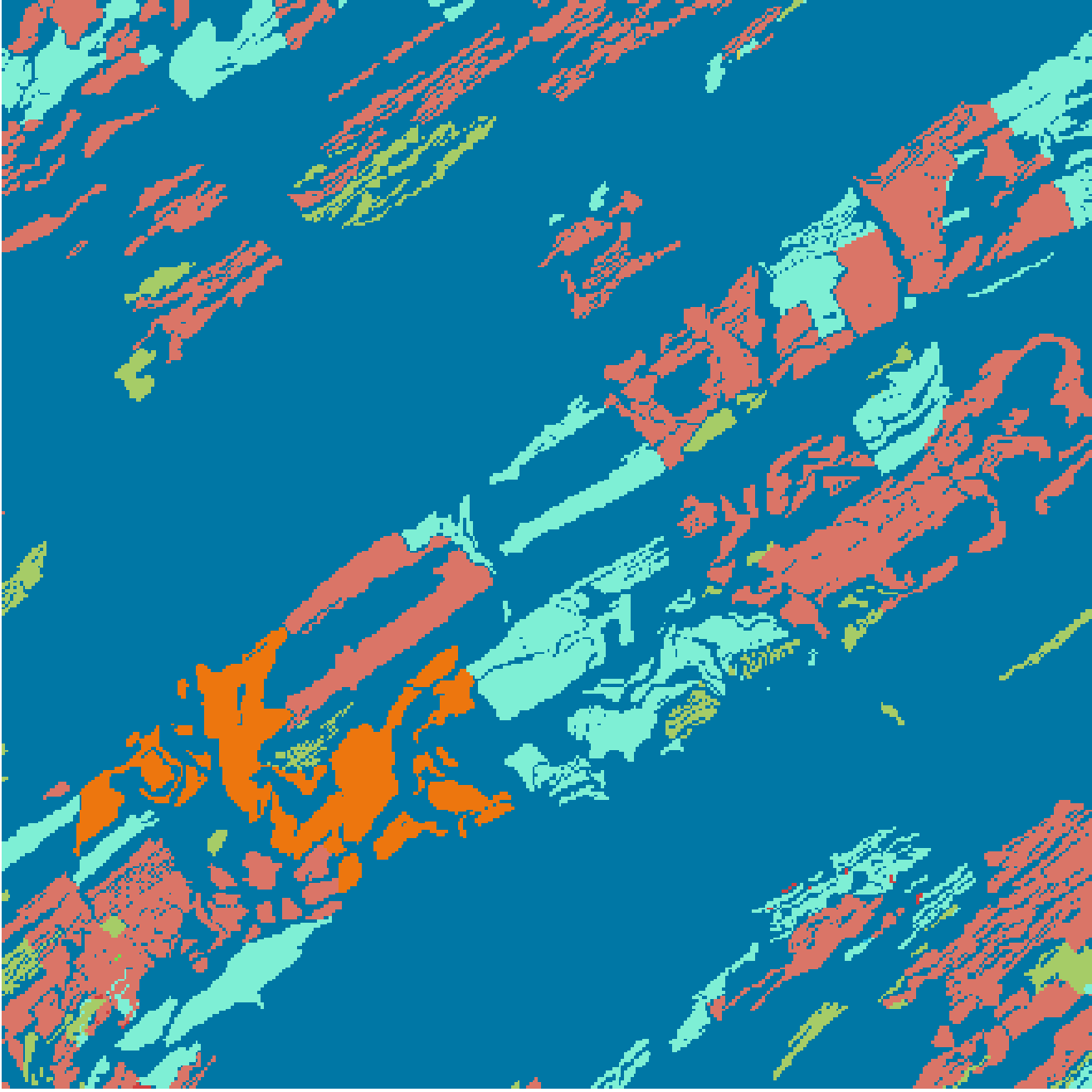}%
    \label{dataset:cat2}}
    \hfil
    \subfloat[]{\includegraphics[width=0.20 \textwidth]{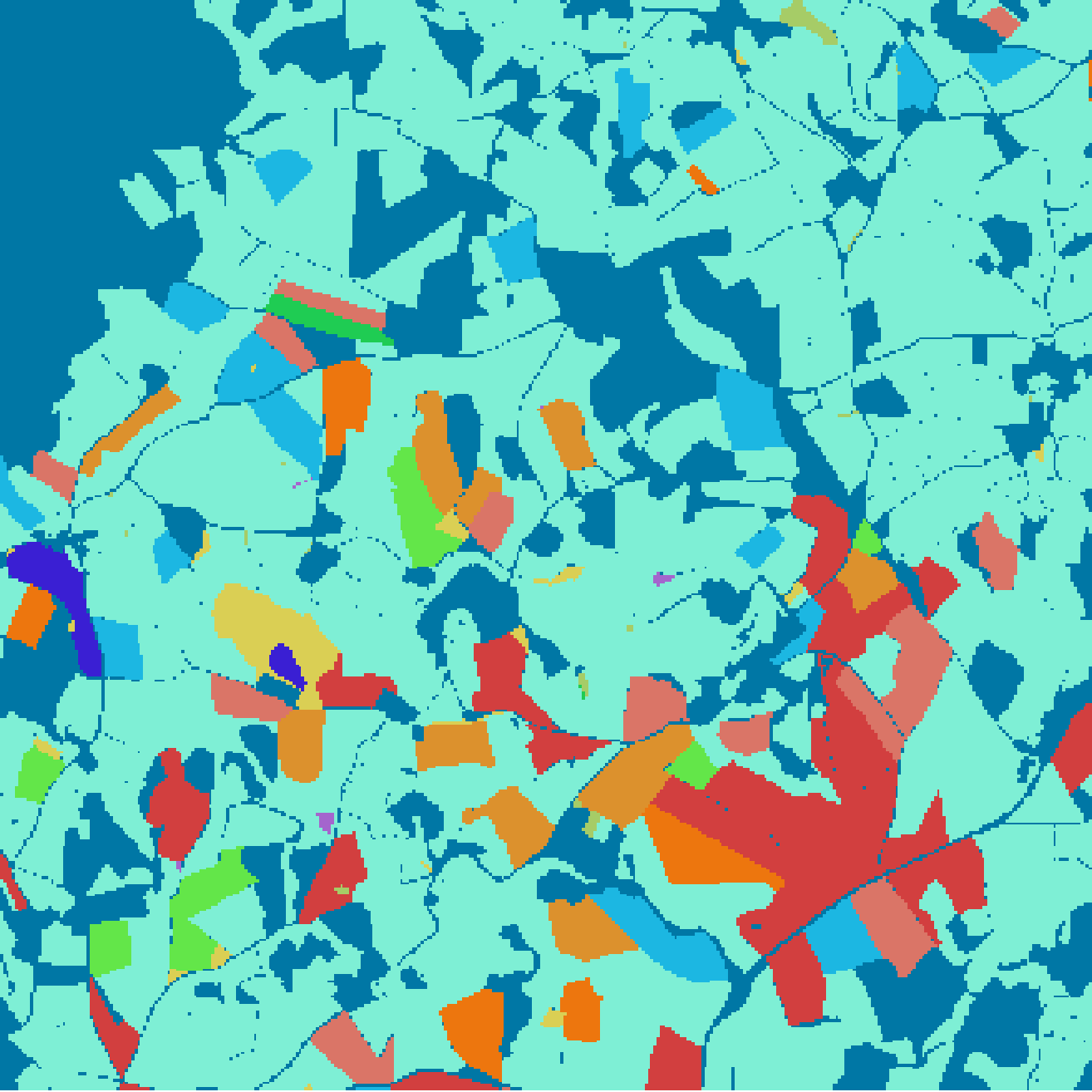}%
    \label{dataset:france1}
    }
    \hfil
    \subfloat[]{\includegraphics[width=0.20 \textwidth]{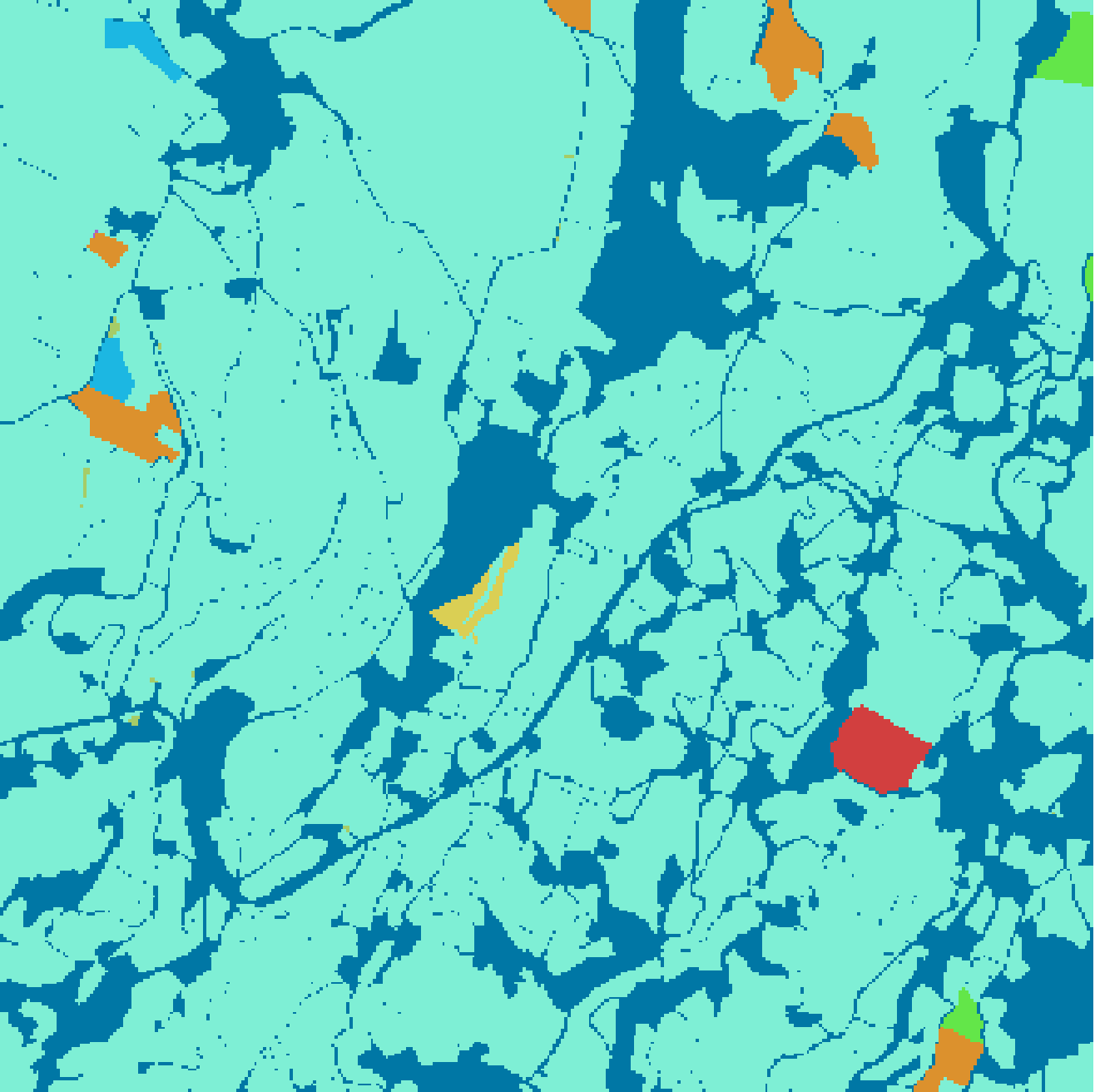}%
    \label{dataset:france2}}
    \hfil
    \subfloat{
    \includegraphics[width=0.1 \textwidth]{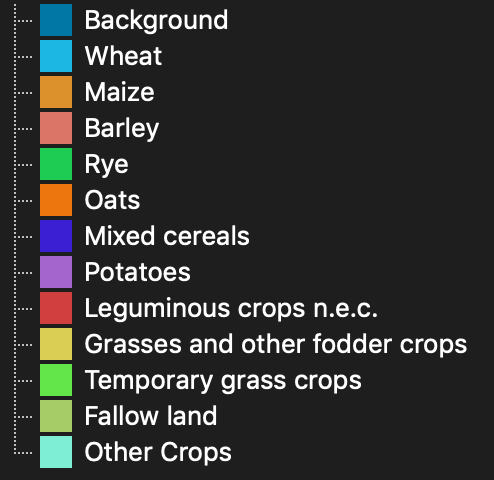}
    }
    
    \caption{Representative patches from the proposed dataset. (a) Catalonia 2019 (31TCG), (b) Catalonia 2020 (31TCG), (c) France 2019 (31TDK), (d) France 2019 (31TDK). Note that (a) and (b) refer to the same patch on different years, whereas (c) and (d) refer to different patches from the same year and region.}
    \label{dataset:visualization}
\end{figure*}

Fig.~\ref{dataset:visualization} presents an example of the annotated Sen4AgriNet dataset, highlighiting also the spatial and temporal variability of the crops and background classes. The first two patches in Fig.~\ref{dataset:visualization} depict the same patch at different years (2019 and 2020), while the next two patches are different patches for the same year and region. The \textit{background} and \textit{other crops} classes in the taxonomy are distinguished on purpose. \textit{Background} refers to any non crop-related label, while \textit{other crops} is a label representing crops that either can not be matched to the existing taxonomy or are unknown. A more detailed view with visual examples about how data loading works can be found at \cite{s4a_dataloading}. Please note that the aforementioned code repository explains both how instances of the same patch transform through time and how empty-data months are treated. Additionally, code examples are provided to assist individuals with custom logic writing.

\subsection{Dataset Structure}

The core Sen4AgriNet dataset is used to generate two sub-datasets, the Object Aggregated Dataset (OAD), and the Patches Assembled Dataset (PAD). These sub-datasets are reduced in size to allow the training of DL models without worrying about the availability of computing resources becoming a bottleneck.

We first construct the PAD sub-dataset, and then, based on this, we subsequently build the OAD sub-dataset. PAD was built using an automated procedure that downloads and processes Sentinel-2 images. This process is tile oriented per available year of LPIS data. Overall, the available LPIS data from France and Catalonia extend 55 Sentinel-2 tiles. All Sentinel-2 L1C images with less than 10\% cloud coverage are selected for download and each image is split into 900 non-overlapping patches. A single patch contains 366x366 images for the 10-meter bands, 183x183 for the 20-meter bands and 61x61 for the 60-meter bands. The size of the patches was chosen in order to have integer division of the size of the tile with all 3 different spatial resolutions of Sentinel-2.

As a next step, the patches that correspond to the same location but have different dates are stacked into a single netCDF file. The netCDF format was chosen because it is self-describing (compared to other formats like geotiff, jpeg2000, etc.), portable, flexible, and is a popular standard for storing geospatial data. Sen4AgriNet consists of thousands of patches/files, thus managing this huge volume of data is a major concern. Furthermore, since we run many different experiments on Sen4AgriNet dataset, the self-descriptiveness flexibility of netCDF allows each experiment to be documented within the corresponding experimental dataset, without the need for auxiliary or explicit documentation tools.

After creating the patches, the labels need to be imported. For each patch the corresponding 366x366 LPIS with the Sen4AgriNet taxonomy applied was rasterized to match the specific pixels in that patch. Using this approach each patch includes a total of $13 * n_{time} + 2$ variables in the netCDF file. Each ``spectral'' variable contains all the $n_{time}$ data acquisitions (depending on the region $n_{time}$ varies from 30 to 50 timestamps for one year). The remaining two variables contain the rasterized LPIS masks. The first mask is a mask of integers corresponding to the crop codes of the taxonomy. The second mask is a 64-bit integer that contains the parcels code number and is mainly useful for individual parcel extraction problems. 

We must note here that the size of the resulting dataset is huge. A single Sentinel-2 tile stack that is transformed into patches for a one year span, results in $\sim 900$ patches in total and has volume ranging between 6 and 44 GBs.

The OAD part is built on top of PAD by aggregating to the parcel level. The pixels contained within each individual parcel are aggregated via the mean function to a single number. This is repeated for each available timestamp within the netCDF variable. The final result is a CSV file containing the individual parcels as rows and the aggregated timestamps for spectral bands as columns.

\begin{figure}[htb]
    \centering
    \includegraphics[width=0.45\textwidth]{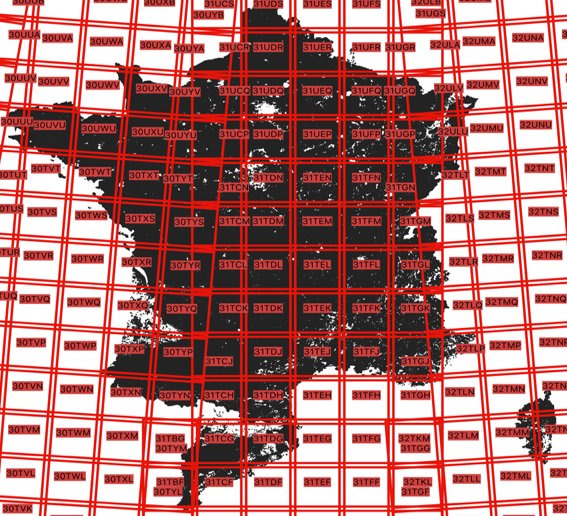}
    \caption{France and Catalonia (Spain) LPIS dataset overlaid with Sentinel-2 tiles.}
    \label{fig:lpis-dataset}
\end{figure}

France’s and Catalonia’s parcels are mainly present in 55 Sentinel-2 tiles (Fig. \ref{fig:lpis-dataset}). The total size of the Sen4AgriNet dataset is estimated to be 10 TB for the 2016-2020 time span, which corresponds to a total of 225,000 patches. 

\vspace{-\baselineskip}

\begin{figure}[h]
    \centering
    \hspace*{-1.1cm}
    \includegraphics[scale=0.36]{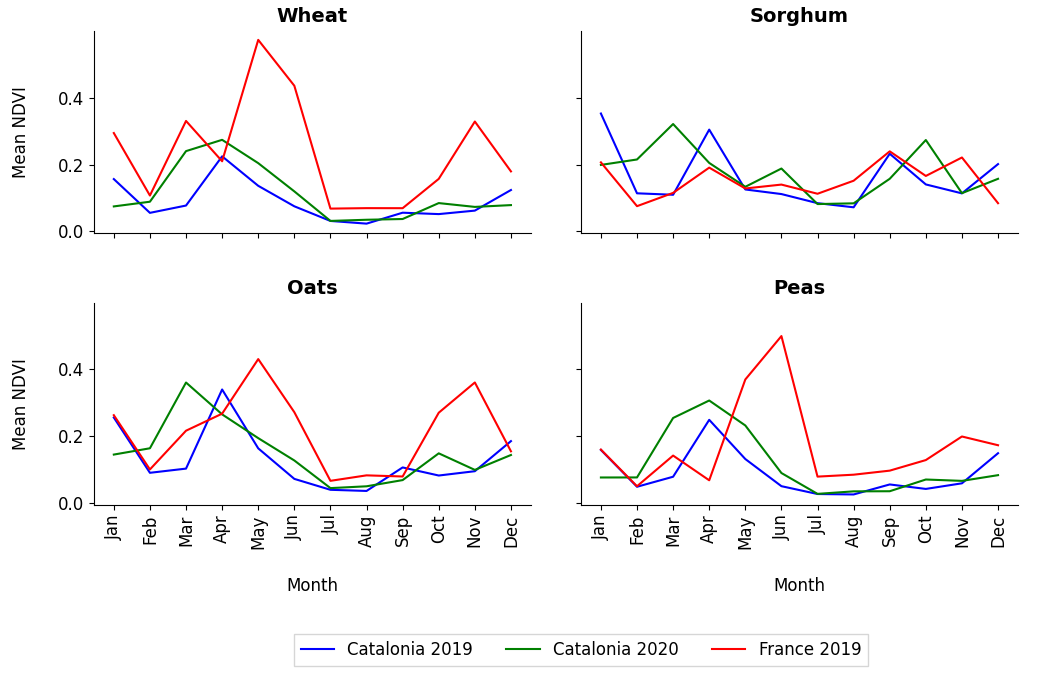}
    \caption{Mean Normalized Difference Vegetation Index (NDVI) of different crops across space and time.}
    \label{ndvis}
\end{figure}
    \section{Experiments}

\begin{figure*}[ht]
    \centering
    \hspace*{-1cm}
    \subfloat[]{\includegraphics[width=1.1 \textwidth]{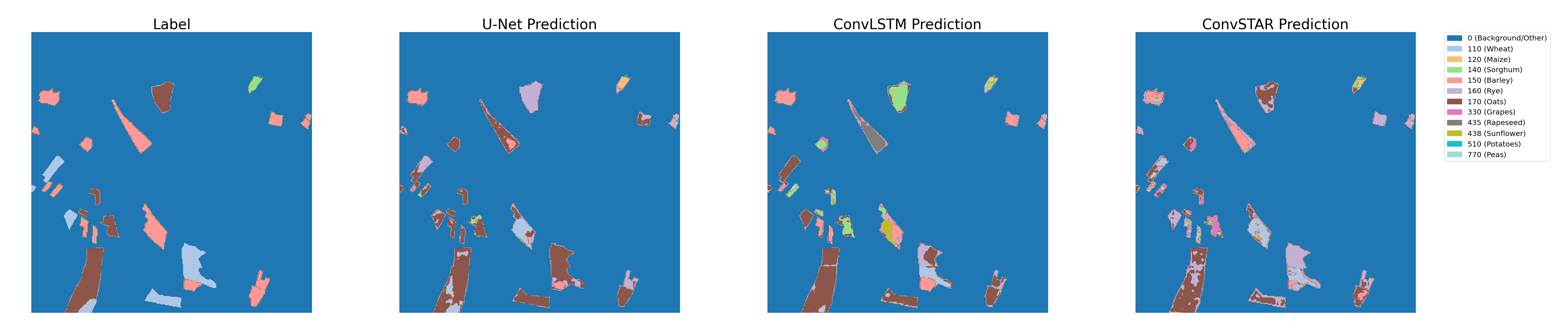}%
    \label{scenario1:img1_convlstm}
    }
    \hfil
    \hspace*{-1cm}
    \subfloat[]{\includegraphics[width=1.1 \textwidth]{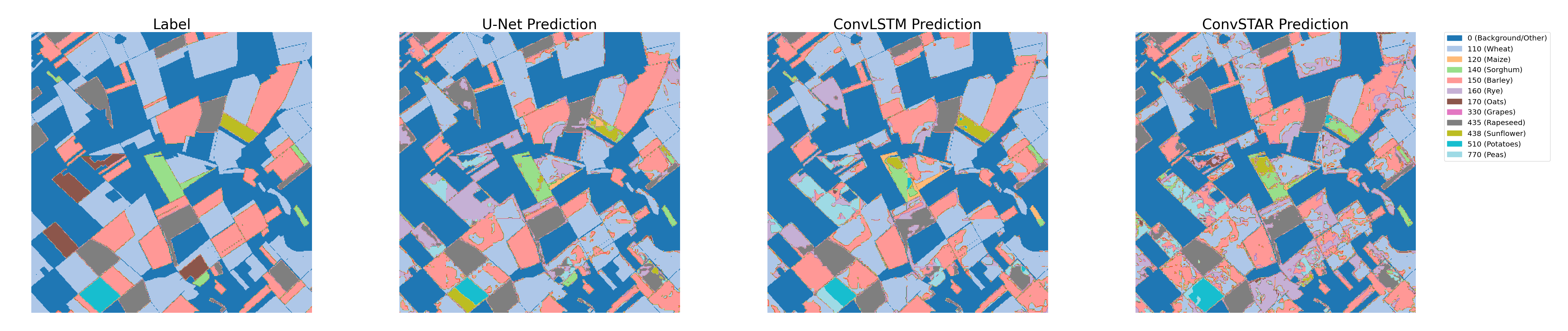}%
    \label{scenario1:img2_convlstm}}
    \caption{Scenario 1. Visual evaluation of the U-Net, ConvLSTM and ConvSTAR PAD predictions for two different image patches.}
    \label{scenario1:vis_convlstm_pad}
\end{figure*}

\begin{figure*}[ht]
    \centering
    \hspace*{-1cm}
    \subfloat[]{\includegraphics[width=1.1 \textwidth]{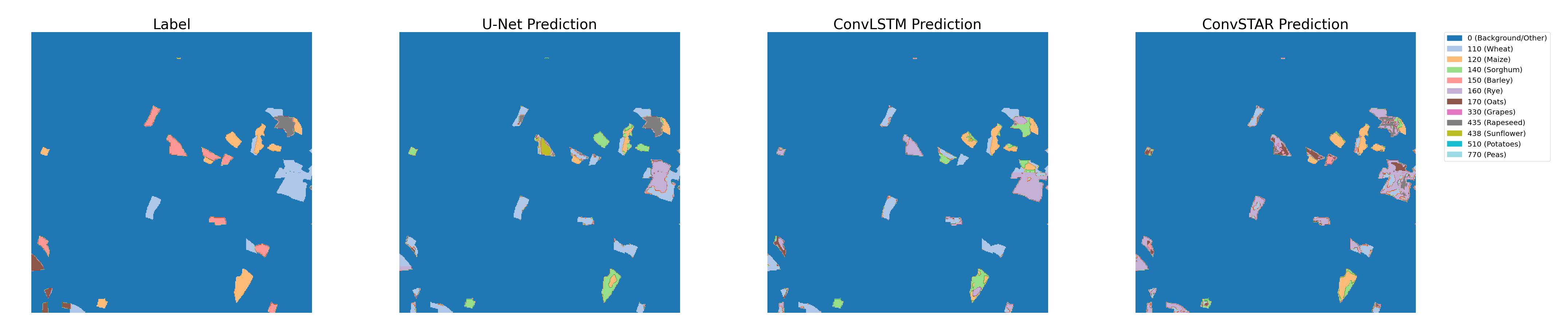}%
    \label{scenario2:img1_convlstm}
    }
    \hfil
    \hspace*{-1cm}
    \subfloat[]{\includegraphics[width=1.1 \textwidth]{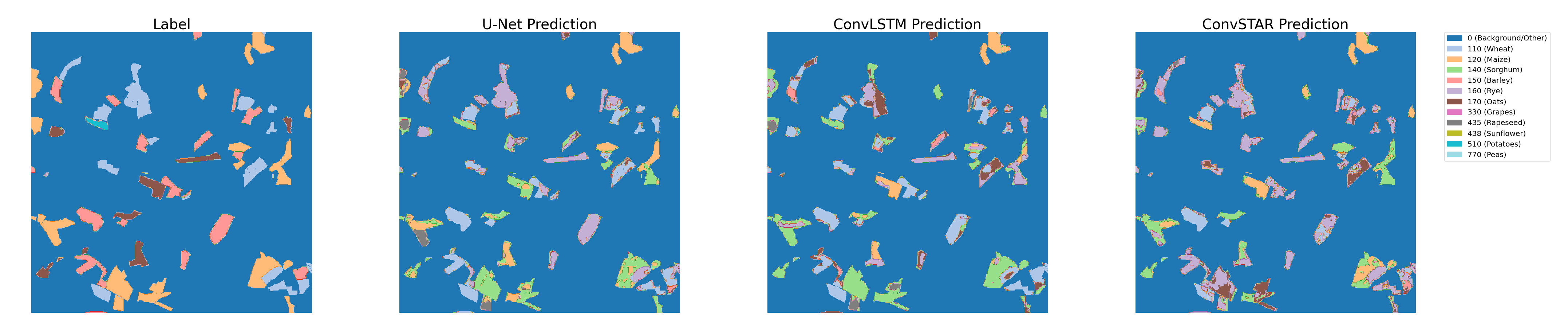}%
    \label{scenario2:img2_convlstm}}
    \caption{Scenario 2. Visual evaluation of the U-Net, ConvLSTM and ConvSTAR predictions for two different image patches.}
    \label{scenario2:vis_convlstm_pad}
\end{figure*}

\begin{figure*}[ht]
    \centering
    \hspace*{-1cm}
    \subfloat[]{\includegraphics[width=1.1 \textwidth]{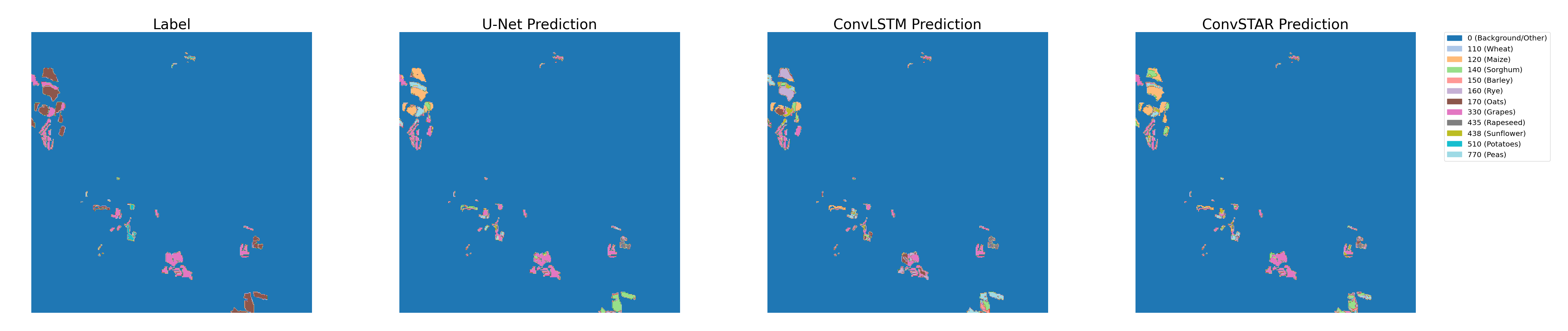}%
    \label{scenario3:img1_convlstm}
    }
    \hfil
    \hspace*{-1cm}
    \subfloat[]{\includegraphics[width=1.1 \textwidth]{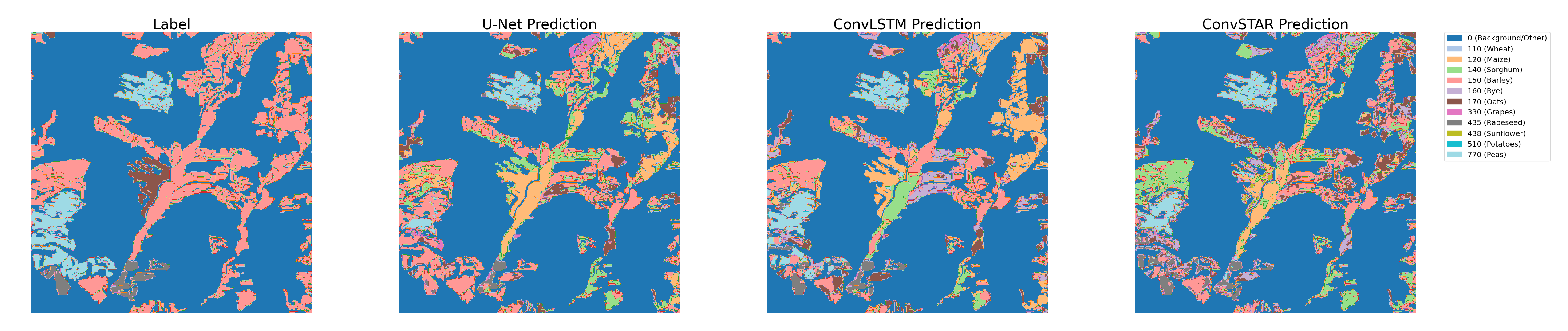}%
    \label{scenario3:img2_convlstm}}
    \caption{Scenario 3. Visual evaluation of the U-Net, ConvLSTM and ConvSTAR predictions for two different image patches.}
    \label{scenario3:vis_convlstm_pad}
\end{figure*}

In this work we extend the experiments presented in~\cite{9553603} by using new DL architectures and new sub-datasets. We expand the geographic and temporal coverage beyond using Catalonia’s part of the Sen4AgriNet dataset for the year 2020~\cite{9553603}, by also including years 2019 and 2020 for the entire region of Catalonia and part of France. We sample 5,000 patches from the entire dataset with label stratification, resulting in a reduced dataset of size $\sim 140$ GB. In order to limit the timesteps for the input time series, we aggregated the data into 12 time-bins by calculating the median of each month, and used a fixed window including the medians of months 4 (April) through 9 (September) for model training. Only bands Red, Green, Blue and Near-Infrared were selected due to their higher spatial resolution.

As a second step, the number of labels was constrained in order to ensure proper number of available observations for training/validating/testing. The label distributions from each region (pixel-wise) were extracted. The common labels between regions were isolated, the cumulative sum of the labels was computed and then all common labels forming the 99.9\% of the data were selected. This resulted in 11 classes (from a 168 total): wheat, maize, sorghum, barley, rye, oats, grapes, rapeseed, sunflower, potatoes and peas. Based on this, the train-validation-test ratio is fixed to 60\%-20\%-20\% for all experiments.

The spatiotemporal variability of the dataset is evident if we examine the observed mean Normalized Difference Vegetation Index (NDVI) of the selected crops. Fig. \ref{ndvis} showcases how the mean NDVI changes over time for wheat, sorghum, oats and peas across the different years and regions. We can see that in Catalonia a similar trend emerges for both years (albeit somehow shifted in time), whereas in France a significantly different crop behaviour is observed. This deviation is justified by both the different climatic conditions and agricultural practices between the different regions. However, crop classification is primarily based on the recognition of the unique growth patterns of the crops, therefore such divergences pose a serious challenge on the design of a robust classifier.

The goal of these experiments is to train DL architectures for crop type classification (OAD) and cropland mapping (PAD) to shed some light into the effects of temporal and spatial crop variability on our models. Our approach is to create challenging scenarios and test the spatio-temporal generalisation performance of state-of-the-art DL models. In Table~\ref{table:experiments} are the three basic scenarios we have identified:
\begin{itemize}
    \item Scenario 1: We use all 5,000 patches together, and create a random split. Data from Catalonia 2019 and 2020, and from France in 2019 are used.
    \item Scenario 2 (spatial generalisation): We use the patches from Catalonia 2019 and 2020 for training, and test on the patches from France 2019.
    \item Scenario 3 (spatio-temporal generalisation): We use the patches from France 2019 for training and test on the patches from Catalonia 2020.
\end{itemize}

\begin{table}[htb]
    \caption{List of scenarios applied in the experiments for PAD and OAD.}
    \label{table:experiments}
    \centering
    \begin{tabular}{c|c|c}
        Scenario & Train & Test\\
        \hline
        1 & \makecell{Catalonia (2019, 2020) \\ France (2019)} & \makecell{Catalonia (2019, 2020) \\ France (2019)} \\
        \hline
        2 & Catalonia (2019, 2020) & France (2019) \\
        \hline
        3 & France (2019) & Catalonia (2020) \\
    \end{tabular}
\end{table}

Due to the different representation of the parcels in the two sub-datasets, different approaches need to be followed in each case. The first approach is based on PAD and considers the labels as rasterized pixel-based maps rather than polygons. Therefore, a semantic segmentation pipeline is employed and detailed pixel-level crop type maps are extracted. By further including the geometry of the parcels additional problems can be defined, such as parcel extraction, parcel counting, etc.  

The second approach, based on OAD, considers the parcels as objects and the raster time series are aggregated per object via zonal statistics (mean, standard deviation, skewness, etc.), thus providing parcel-based insights to the model. This proves rather useful when the parcel geometries are already known (e.g. farmers declarations from EU paying agencies) and the need to simultaneously identify parcels and crop types becomes obsolete. Of course, when the initial geometries are not available then PAD trained models are the solution. Finally, OAD trained models can be used as input to the PAD dataset if the specific DL architecture utilizes pixel based input (e.g. LSTM).

For the evaluation of our experiments, we use the accuracy, precision and F1 scores. Due to high class imbalance, we weigh all metrics with the samples of each class. Also, the corresponding confusion matrices of each experiment and scenario are plotted, normalized across predictions.

    \subsection{Sen4AgriNet – PAD Experiment}

In the first set of experiments, we employed three popular models for image segmentation: U-Net~\cite{unet}, ConvLSTM~\cite{shi_convolutional_2015} and ConvSTAR~\cite{TURKOGLU2021112603}. First, a 3-layer U-Net ($\sim 1.9$m trainable parameters) with LogSoftmax activation and a weighted negative log-likelihood loss was employed as a simple and robust baseline for semantic segmentation. Then, a 3-layer ConvLSTM with an encoder-decoder structure ($\sim 9$m trainable parameters) and LogSoftmax activation at the last layer was trained. Intermediate layers utilize LeakyReLU activations, while the loss function is a weighted negative log-likelihood loss. Second, a 3-layer ConvSTAR model ($\sim 260$k trainable parameters) was also used in comparison, since it is considered more robust and shows more stable convergence. Similar to ConvLSTM, a final LogSoftmax activation was added for the final pixel classification and the loss function of choice was the weighted negative log-likelihood. Due to time constraints both models adopt the architecture proposed in their corresponding publications.

Input patches were divided into non-overlapping sub-patches of size 61x61 for faster computation, so the input is a 6x4x61x61 time series (TxCxHxW, T: timesteps, C: channels, H: height, W: width). Specifically for the U-Net model, images from different timesteps were concatenated along the channels dimension resulting in an input of shape 24x61x61 (T*CxHxW). During training, we masked all non-parcel or unknown pixels and let the models learn only from the known 11 labels. Similarly for the inference stage, all evaluation metrics were calculated on the known parcel labels only. Finally, the Adam optimizer with an initial learning rate of 0.001 was used and a learning rate reduction scheme was employed on validation loss plateaus.

\subsubsection{Scenario 1}

\begin{figure*}[h]
    \centering
    \subfloat[]{\includegraphics[width=0.32 \textwidth]{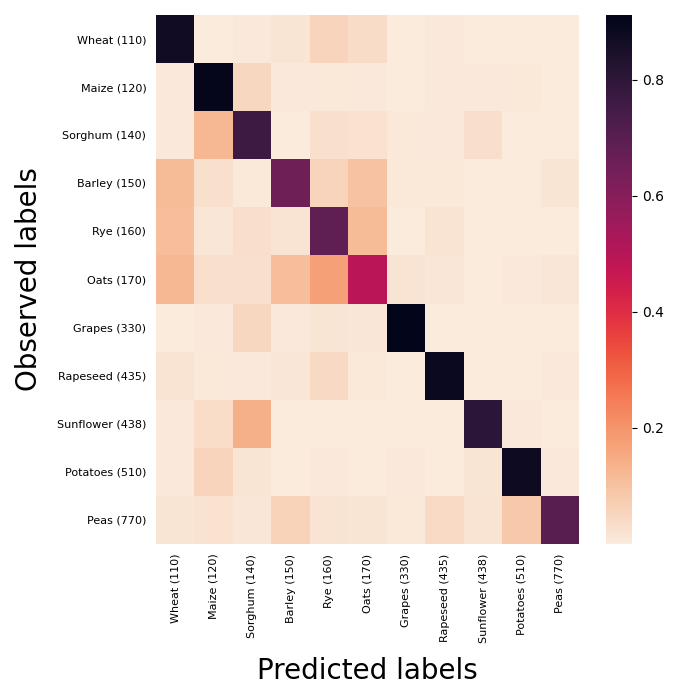}}
    \subfloat[]{\includegraphics[width=0.32 \textwidth]{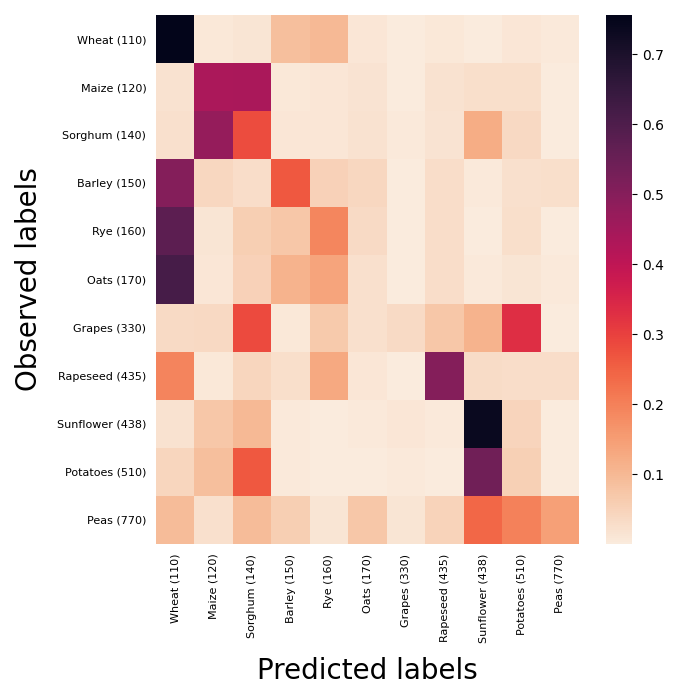}}
    \subfloat[]{\includegraphics[width=0.32 \textwidth]{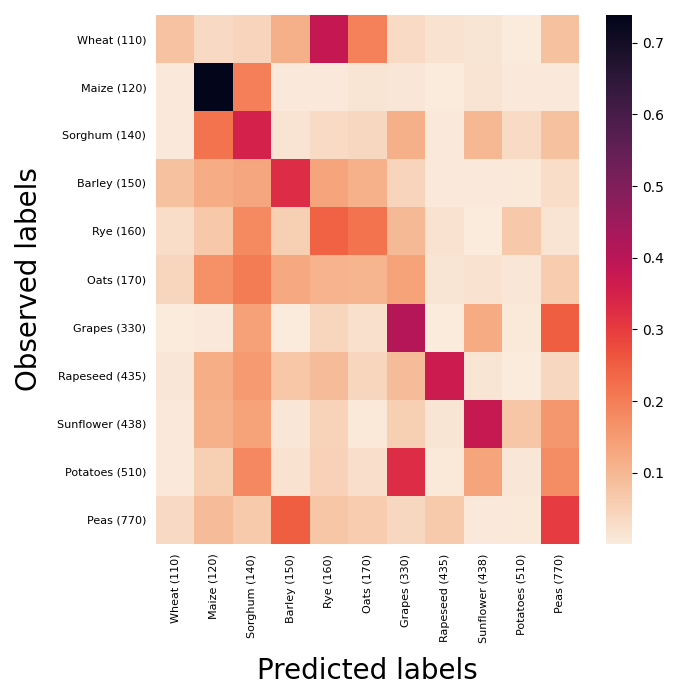}}
    \hfil
    \subfloat[]{\includegraphics[width=0.32 \textwidth]{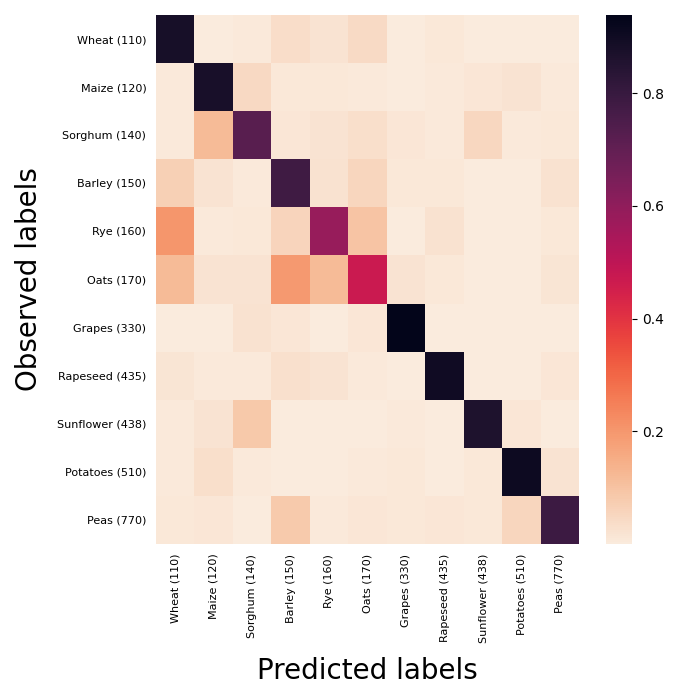}}
    \subfloat[]{\includegraphics[width=0.32 \textwidth]{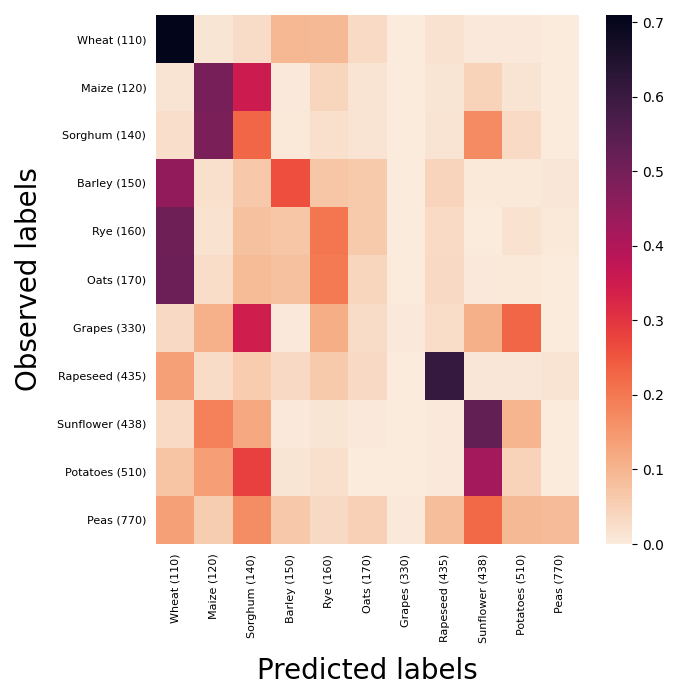}}
    \subfloat[]{\includegraphics[width=0.32 \textwidth]{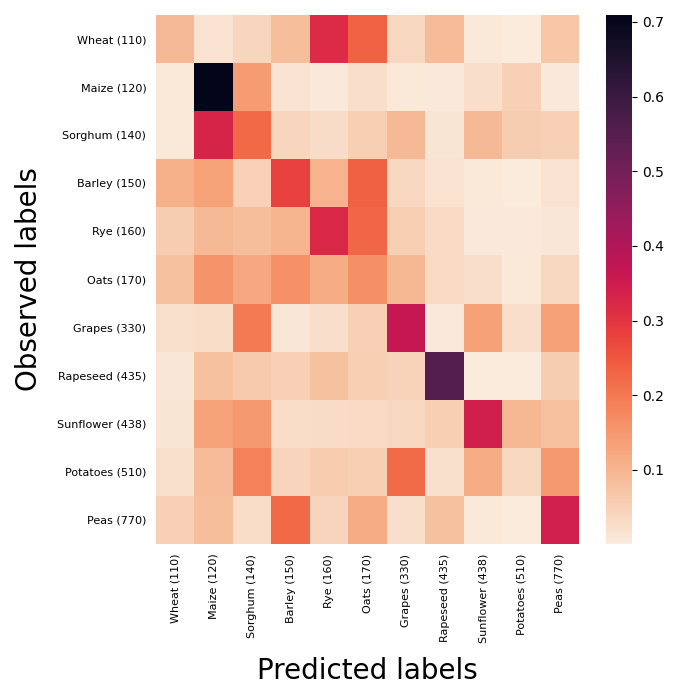}}
    \hfil
    \subfloat[]{\includegraphics[width=0.32 \textwidth]{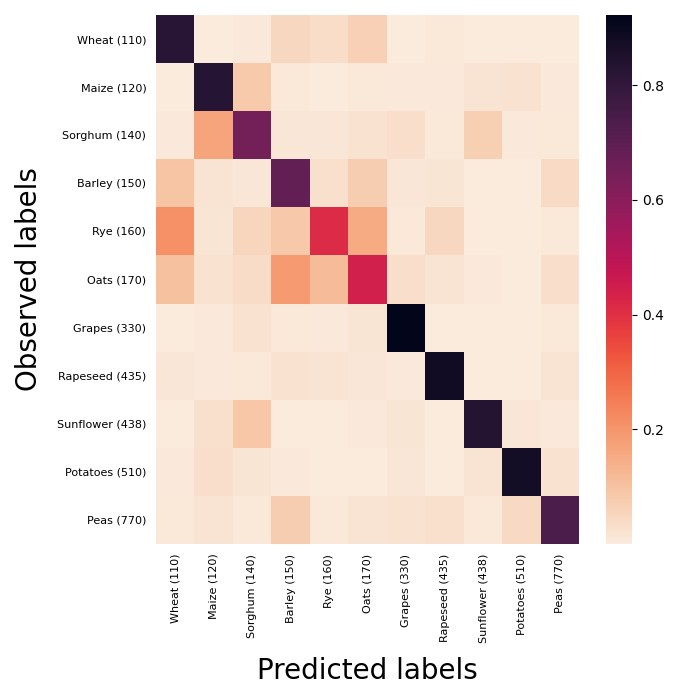}}
    \subfloat[]{\includegraphics[width=0.32 \textwidth]{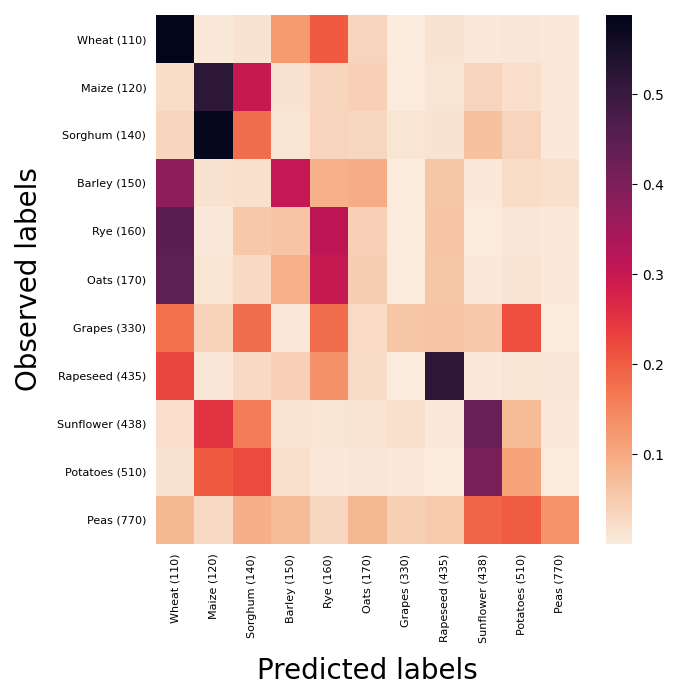}}
    \subfloat[]{\includegraphics[width=0.32 \textwidth]{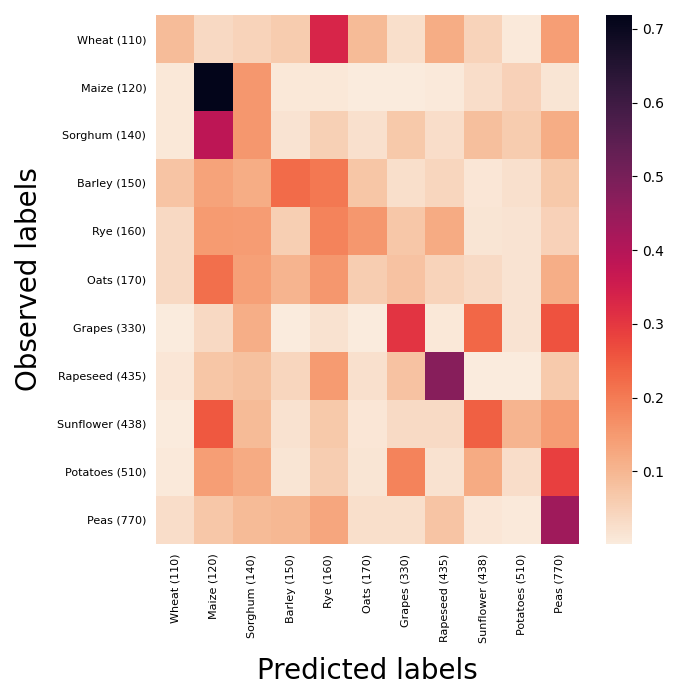}}
    \caption{PAD Results. The first row shows the U-Net results, the second one the ConvLSTM results whereas the third the ConvSTAR results. Each column corresponds to a different scenario.}
    \label{fig:pad-results}
\end{figure*}

In the first scenario, where all regions and years are randomly sampled both in the training and test datasets, ConvLSTM performed significantly better than the other two models in all three summary metrics (Table~\ref{table:pad-scenario1-results}). 
The confusion matrices (Fig.~\ref{fig:pad-results}) and the visual inspection of the predictions (Fig.~\ref{scenario1:vis_convlstm_pad}) support the high valued summary metrics.
An interesting observation from the confusion matrices (Fig.~\ref{fig:pad-results}) is that all models seem to confuse the same crop types, with barley, rye, oats and wheat being the most characteristic case. This is somewhat expected since they are all cereal crops with similar phenology.

\subsubsection{Scenario 2}

In the second scenario, all models were evaluated for their adaptation from one region to another. As shown in the results, all architectures achieve lower accuracy metrics with significantly reduced precision and F1 scores. Again, the corresponding confusion matrices (Fig.~\ref{fig:pad-results}) show similar misclassification patterns between the three models, with wheat, rapeseed and sunflower being the most correctly classified crops. Fig.~\ref{scenario2:vis_convlstm_pad} shows a selection of patches to visually appreciate correct and wrong classifications of the models, with respect to the ground truth.

\subsubsection{Scenario 3}

In the third scenario, all metrics were considerably degraded and all architectures showcased similar performance. From the confusion matrices  (Fig.~\ref{fig:pad-results}), only maize and rapeseed display the highest accuracy. Similarly, in Fig.~\ref{scenario3:vis_convlstm_pad} misclassifications are widespread to all parcels in the selected patches.  

\begin{table}[ht]
    \caption{Results on all scenarios for PAD. Best results are marked in bold.}
    \label{table:pad-scenario1-results}
    \centering
    \begin{tabular}{c|c|c|c|c}
        Scenario & Model & Acc. W (\%) & F1 W. (\%) & Precision W. (\%) \\
        \hline
         & U-Net & 93.70 & 82.61 & 86.64 \\
        1 & ConvLSTM & \textbf{94.72} & \textbf{85.18} & \textbf{86.86} \\
         & ConvSTAR & 92.78 & 80.38 & 83.33 \\
        \hline
         & U-Net & \textbf{83.12} & \textbf{57.85} & \textbf{61.57} \\
        2 & ConvLSTM & 82.53 & 56.56 & 60.57 \\
         & ConvSTAR & 79.52 & 52.15 & 58.98 \\
        \hline
         & U-Net & \textbf{72.11} & \textbf{43.54} & \textbf{68.42} \\
        3 & ConvLSTM & 69.86 & 40.47 & 66.17 \\
         & ConvSTAR & 69.07 & 34.45 & 67.43 \\
    \end{tabular}
\end{table}
    \subsection{Sen4AgriNet - OAD Experiment}

In the object-based set of experiments, three architectures were utilized in order to evaluate model performance for the three different scenarios. The same patches as in the PAD section were selected in order to limit the number of differentiating factors among the two sets of experiments and focus on understanding the generalization performance of the two approaches. 

The first architecture is a network consisting of 3 bidirectional LSTM \cite{lstm} layers with a hidden layer of size 1024, alongside one linear layer (applied to the last hidden state), a ReLU and a final classification linear layer. The entire model includes $\sim 60$m trainable parameters.

Since their first introduction, transformer networks \cite{NIPS2017_3f5ee243} are considerably valued as an extraordinary opponent to the LSTM networks. Thus, the second DL architecture is a transformer-encoder classifier that was designed to be tested against the well known LSTM. Instead of calculating an embedding space, the precomputed parcel statistics are used as input. The dimensionality of the encoder-transformer type network uses a 26 features input (2 statistics for each band channel) and 2 heads, as heads should divide the number of features. For the forward function, a customized Positional Encoder was utilized. This Positional Encoder features the vanilla sine and cosine functions proposed by the authors of \cite{NIPS2017_3f5ee243} for encoding each token position. After encoding the position, the transformer encoder is applied and then a final sequence of two linear layers is used. The final model contains a total of $\sim 270$k trainable parameters.

Other hyperparameters such as the number of encoder layers and the hidden dimension size are kept the same as in the LSTM runs (3 layers, 1024-dimensional hidden size). Both experiments use the Adam optimizer with initial learning rates 0.001 and 0.0001 respectively, as well as a step learning rate scheduler of 0.1 every 5 epochs. The Cross Entropy is used as the loss function.

Finally, a quite famous deep learning architecture in the Remote Sensing field is being marshalled. TempCNN \cite{rs11050523, breizhcrops2020} with a total of $\sim 719$k trainable parameters, where convolution stages are designed to capture the time aspect of the OAD dataset.

Results obtained from model will be compared against each other for every test scenario.

\begin{table}[htb]
    \caption{Results on all scenarios for OAD. Best results are marked in bold.}
    \label{table:oad-scenario1-results}
    \centering
    \begin{tabular}{c|c|c|c|c}
        Scenario & Model & Acc. W (\%) & F1 W. (\%) & Precision W. (\%) \\
        \hline
        1 & LSTM & 88.52 & 88.03 & 87.85 \\
         & Transformer & 88.36 & 88.10 & 87.90 \\
         & TempCNN & \textbf{90.08} & \textbf{89.97} & \textbf{90.01} \\
        \hline
        2 & LSTM & \textbf{91.55} & \textbf{91.34} & \textbf{91.31} \\
         & Transformer & 39.17 & 31.45 & 58.52 \\
         & TempCNN & 36.90 & 30.14 & 60.71 \\
        \hline
        3 & LSTM & \textbf{60.60} & \textbf{63.96} & \textbf{70.55} \\
         & Transformer & 51.21 & 56.71 & 67.76 \\
         & TempCNN & 52.32 & 57.38 & 68.35 \\
    \end{tabular}
\end{table}

\subsubsection{Scenario 1}

\begin{figure*}[ht]
    \centering
    \subfloat[]{\includegraphics[width=0.3 \textwidth]{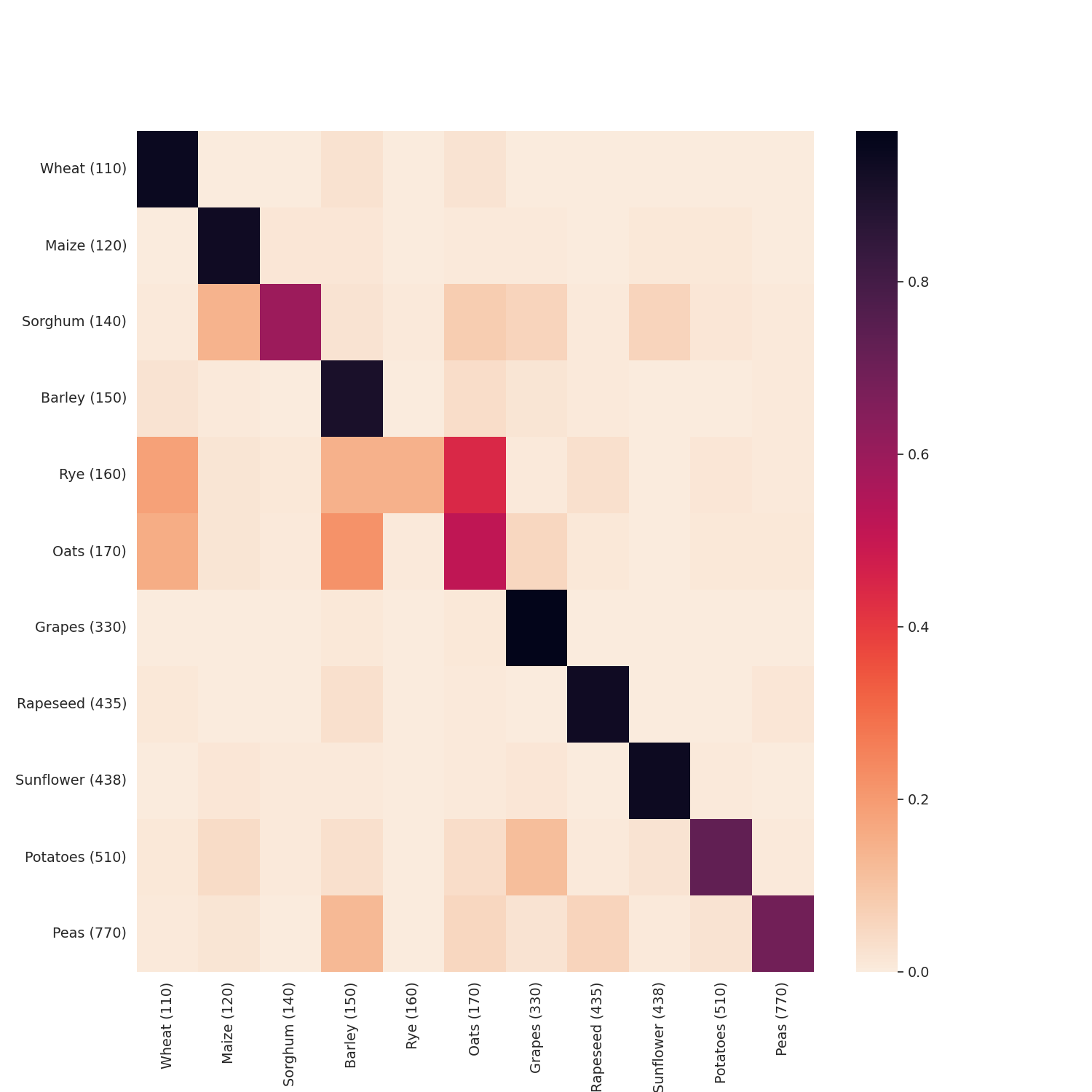}
    \label{fig:oad-results-lstm1}}
    \subfloat[]{\includegraphics[width=0.3 \textwidth]{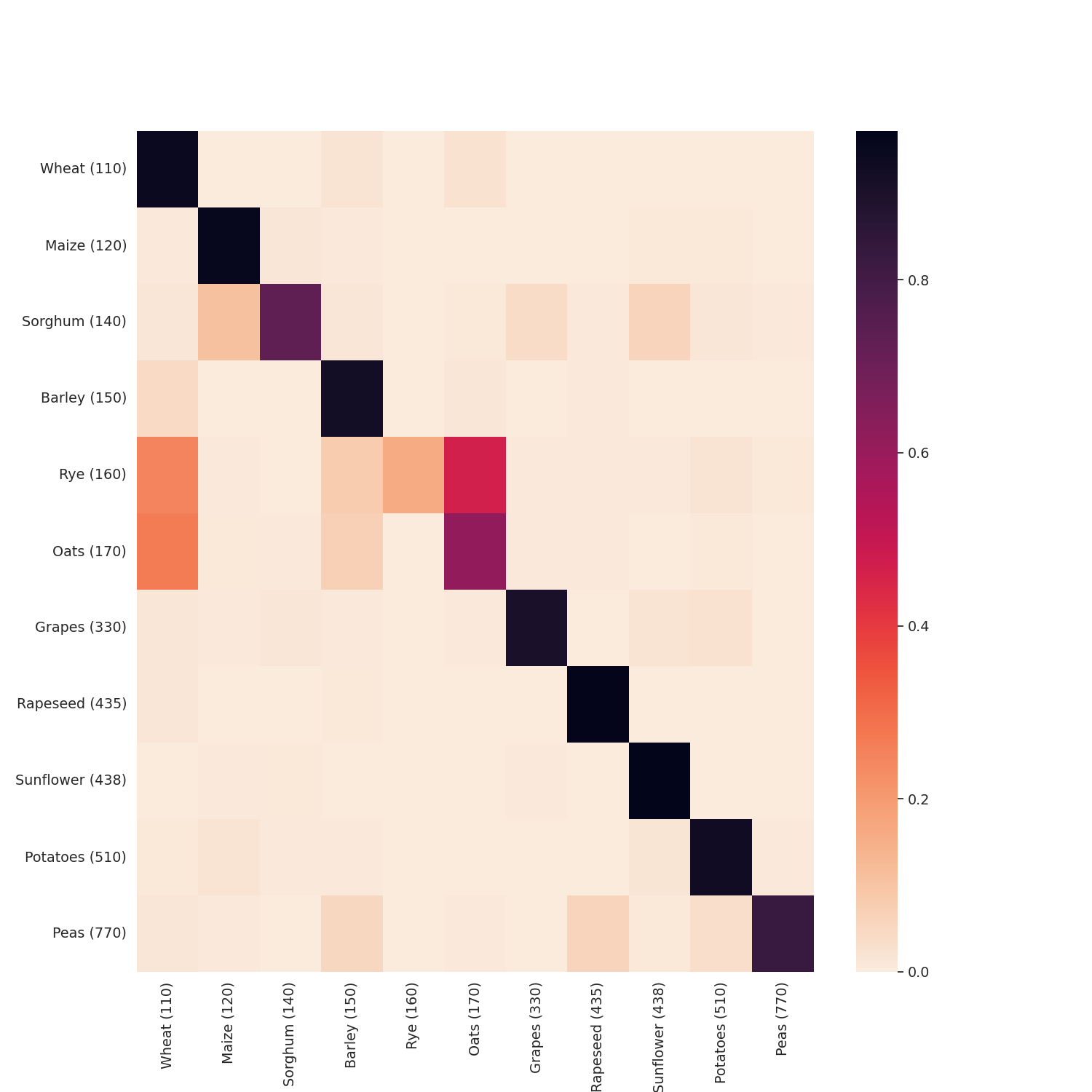}
    \label{fig:oad-results-lstm2}}
    \subfloat[]{\includegraphics[width=0.3 \textwidth]{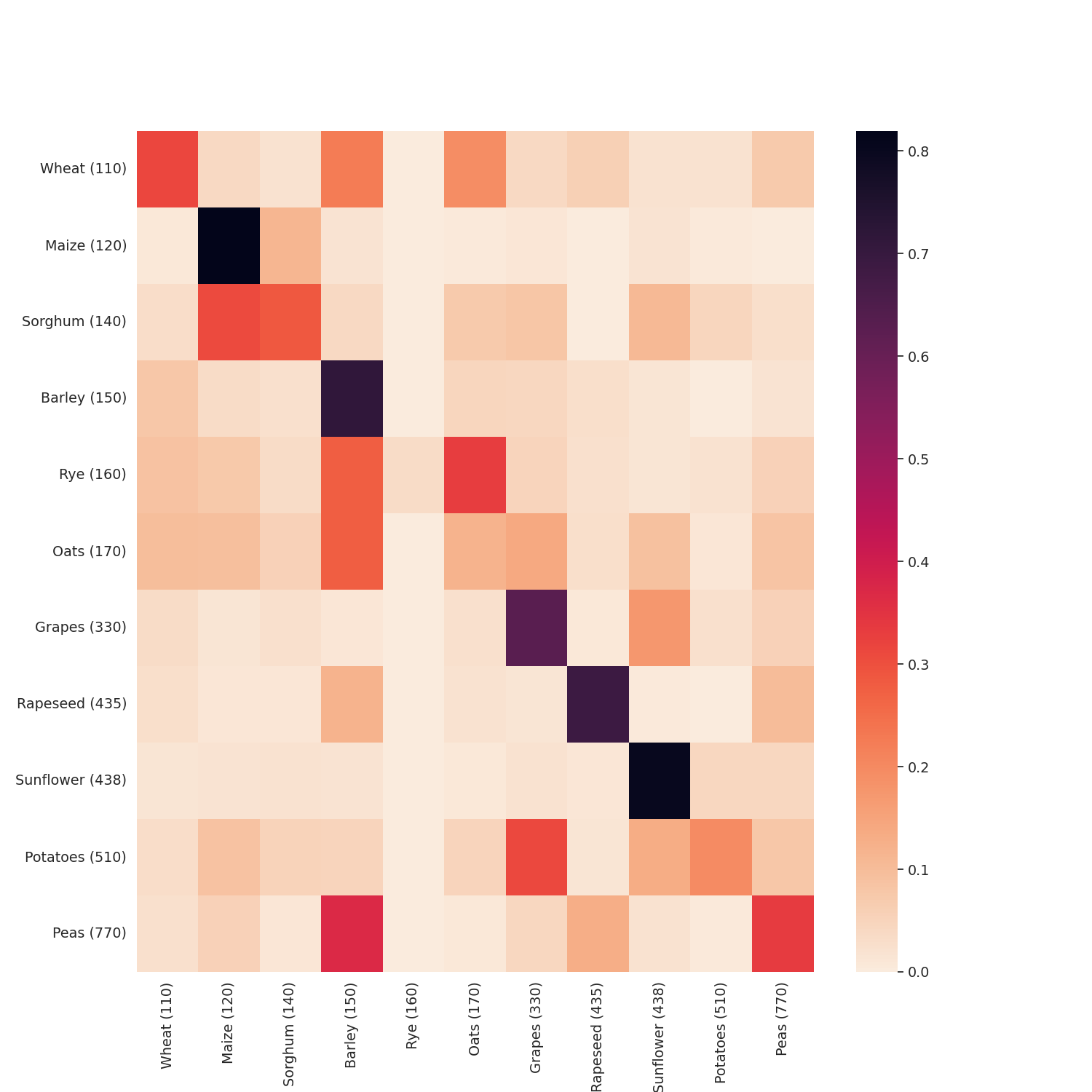}
    \label{fig:oad-results-lstm3}}
    \hfil
    \subfloat[]{\includegraphics[width=0.3 \textwidth]{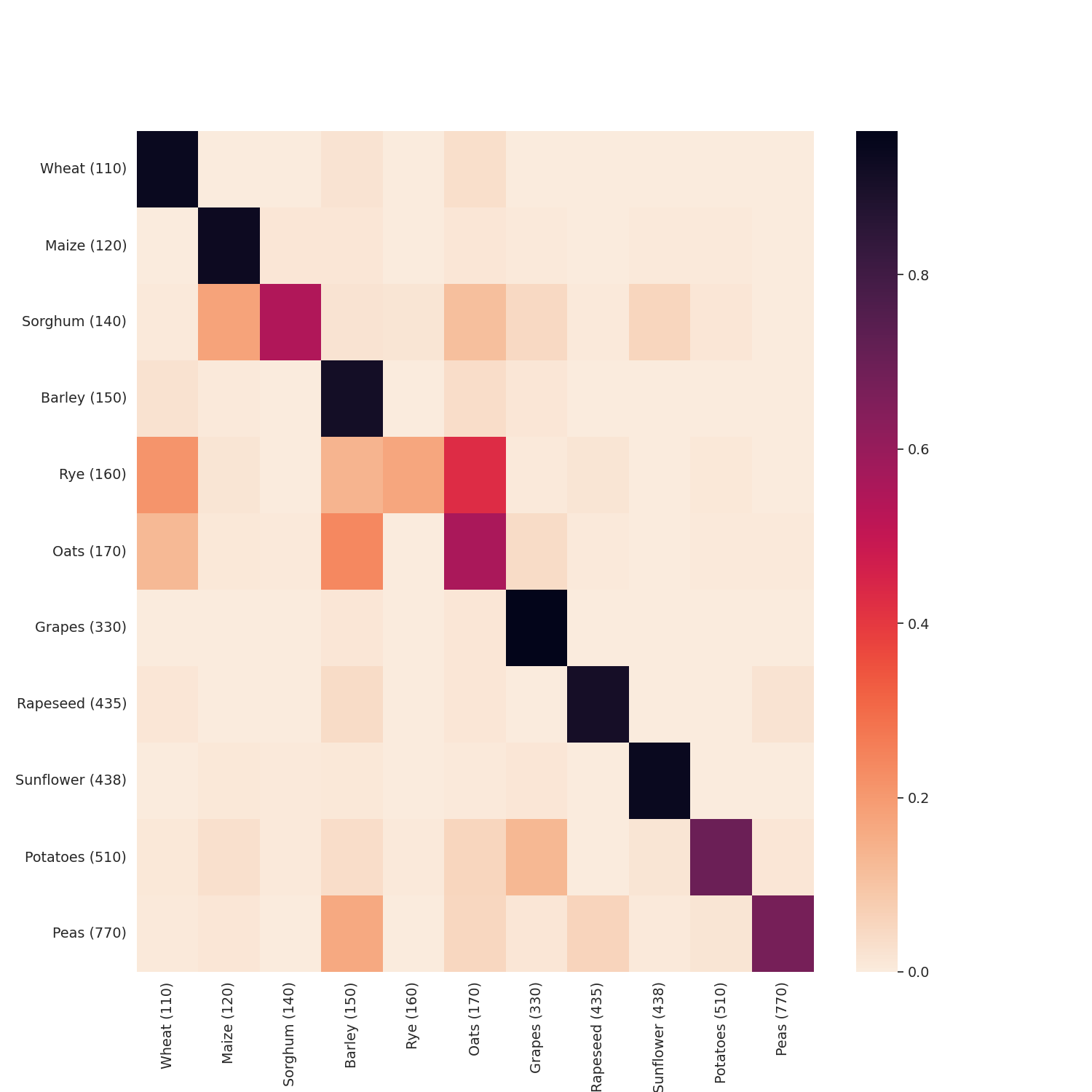}
    \label{fig:oad-results-transformer1}}
    \subfloat[]{\includegraphics[width=0.3 \textwidth]{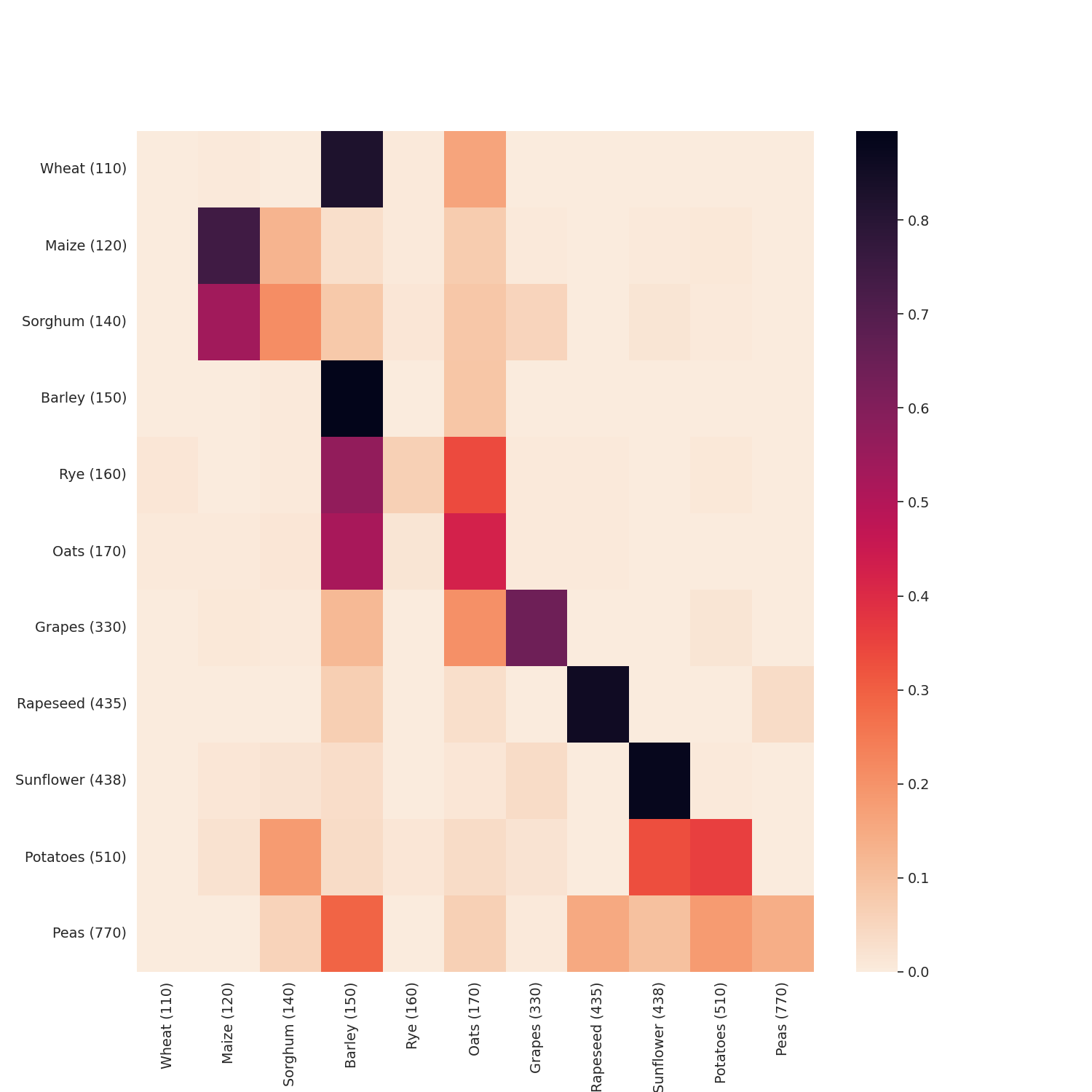}
    \label{fig:oad-results-transformer2}}
    \subfloat[]{\includegraphics[width=0.3 \textwidth]{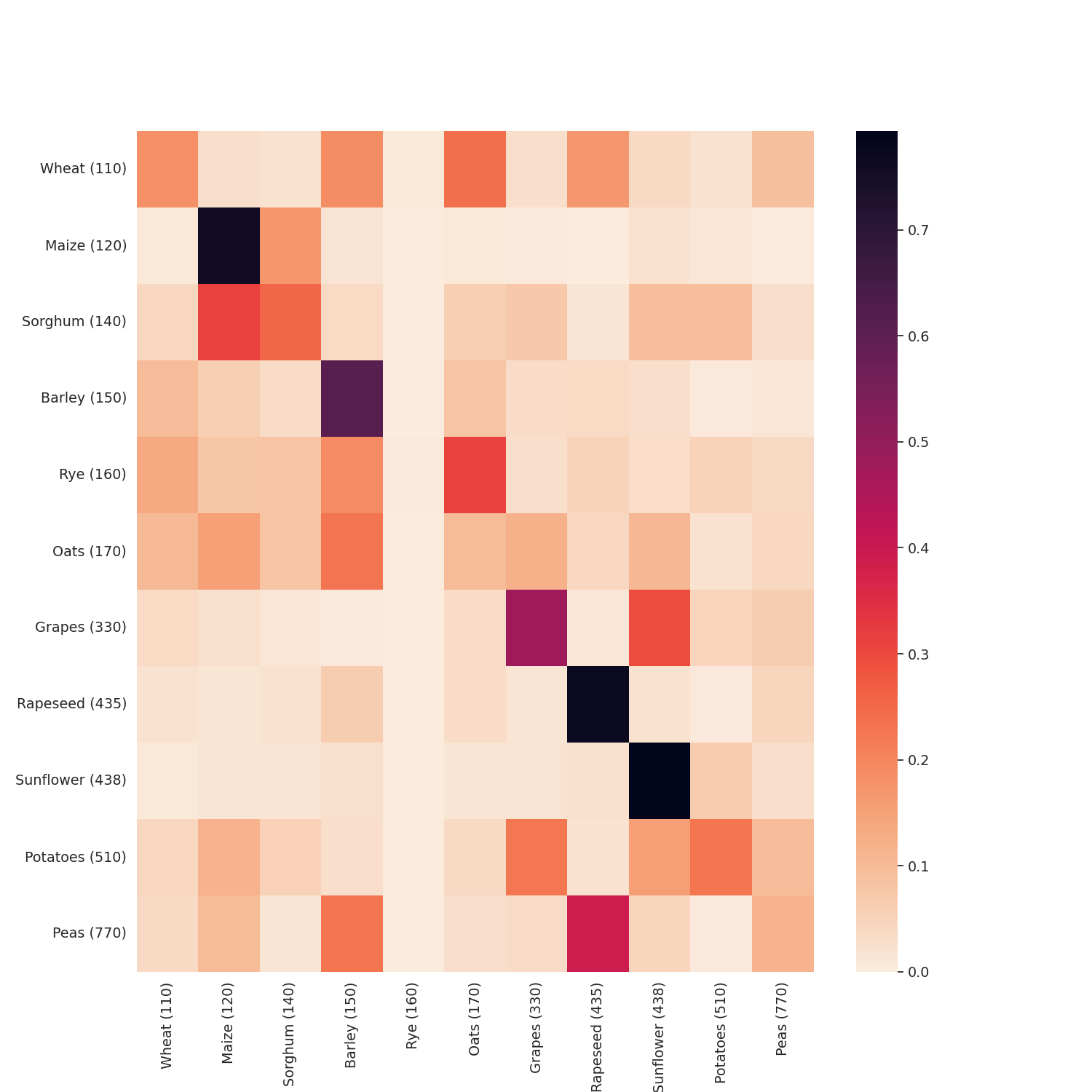}
    \label{fig:oad-results-transformer3}}
    \hfil
    \subfloat[]{\includegraphics[width=0.3 \textwidth]{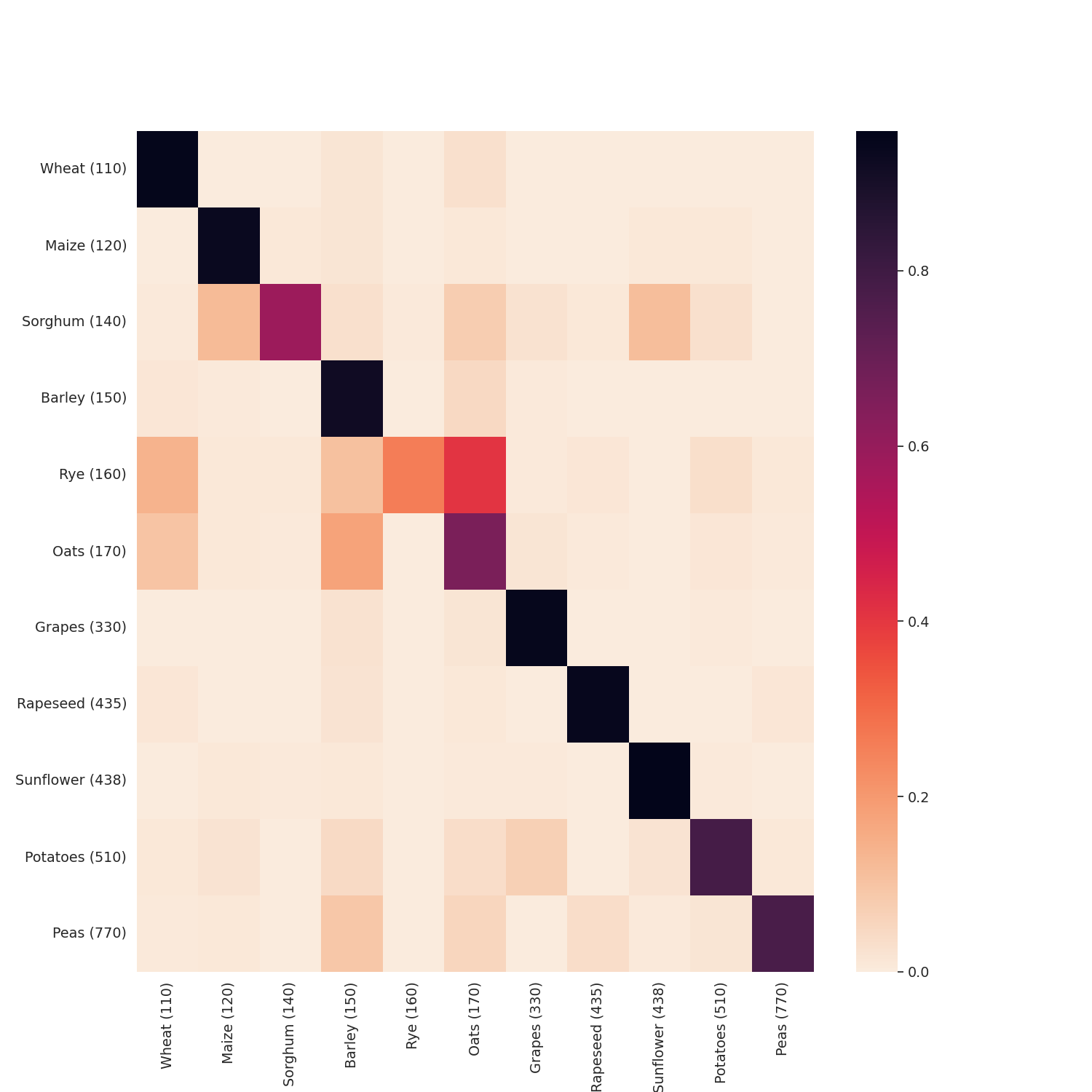}
    \label{fig:oad-results-tempcnn1}}
    \subfloat[]{\includegraphics[width=0.3 \textwidth]{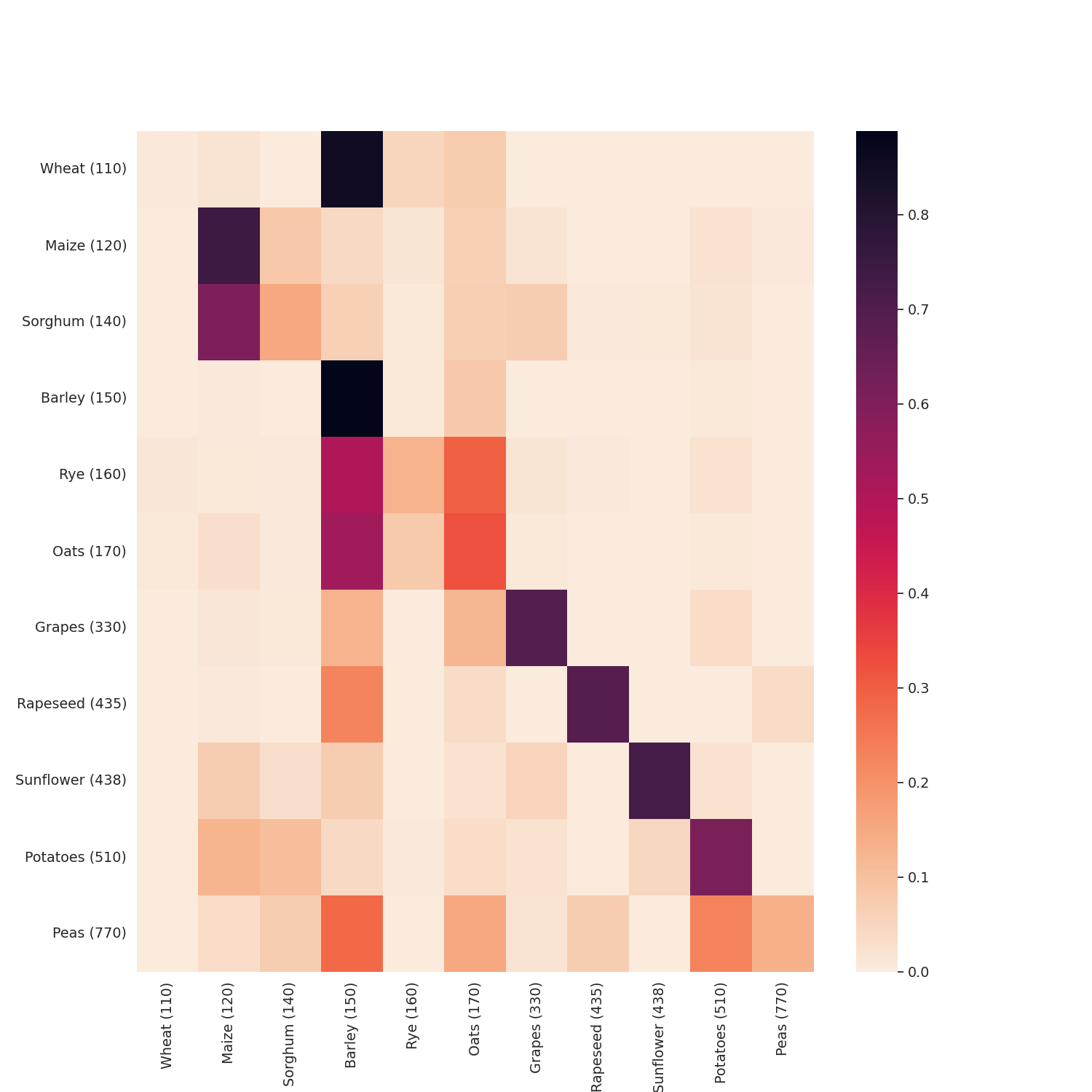}
    \label{fig:oad-results-tempcnn2}}
    \subfloat[]{\includegraphics[width=0.3 \textwidth]{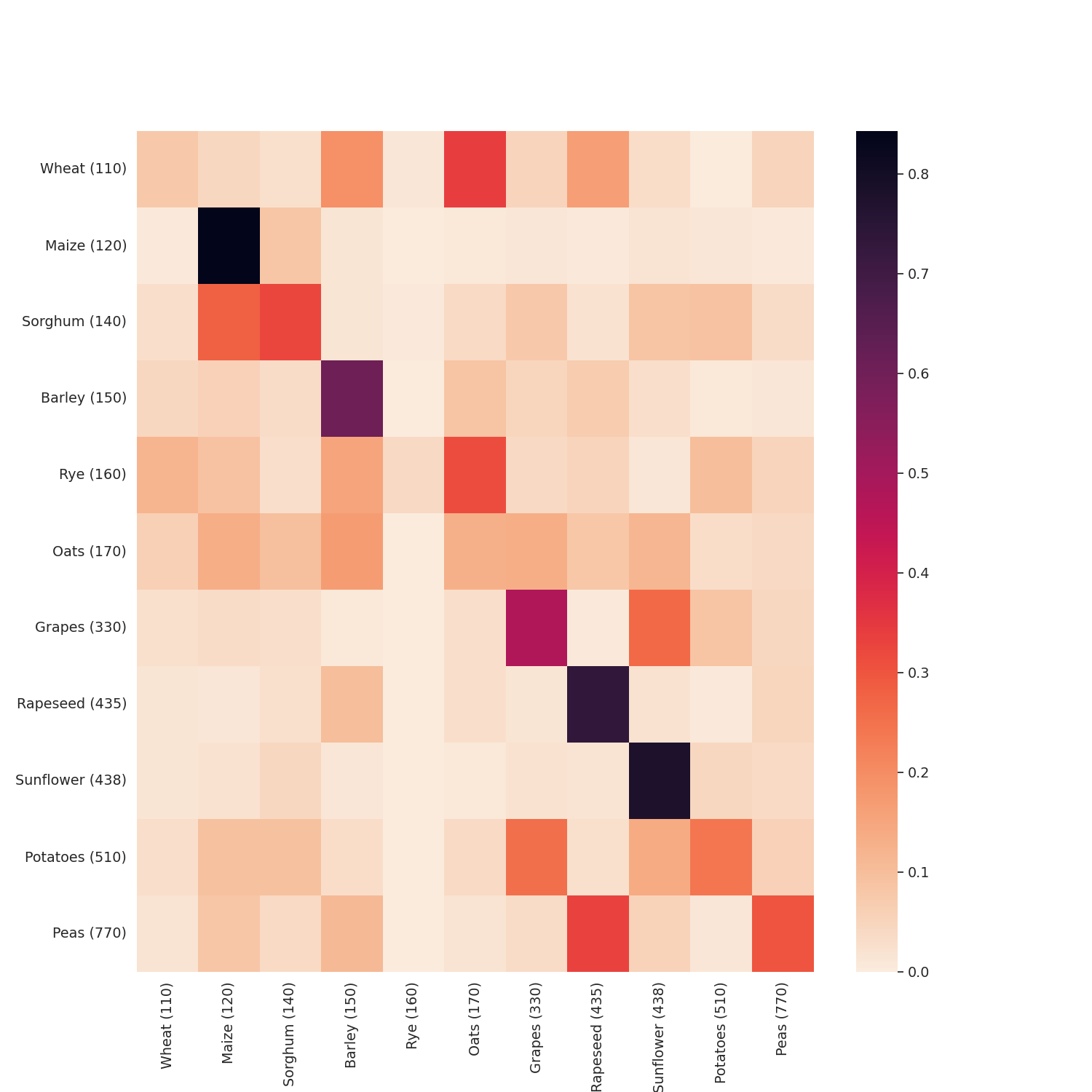}
    \label{fig:oad-results-temp3}}
    \caption{OAD Results. The first row shows the LSTM results whereas the second row the transformer-encoder classifier results. The third row illustrates results obtained for the TempCNN architecture. Each column corresponds to a different scenario.}
    \label{fig:oad-results}
\end{figure*}

In the first scenario, the three different architectures (LSTM, transformer, TempCNN) were evaluated. All architectures displayed quite similar performance as shown in the metrics Table~\ref{table:oad-scenario1-results}, both in the overall classification evaluation and in the individual classes (confusion matrices). The TempCNN architecture achieves slightly better results. As seen in the corresponding confusion matrices (Fig.~\ref{fig:oad-results-lstm1}, \ref{fig:oad-results-transformer1} and \ref{fig:oad-results-tempcnn1}) most of the crop type categories are classified correctly. Though, a specific misclassification pattern among some classes is observed in both architectures. Rye is mostly confused with wheat, barley and oats, while oats are confused with rye, barley and wheat. On the contrary, wheat is hardly confused with other classes. This behaviour can be firstly attributed to the fact that wheat, rye, barley and oats are crops with very similar spectral and temporal signatures. Secondly, wheat is a dominant class (39.4\%), while the other three classes together account for 27.6\% of the training data.

\subsubsection{Scenario 2}

In the second scenario, the training was performed on Catalonia for both available years and tested on France 2019 labeled data. Impressively, the LSTM model significantly outperformed both the TempCNN and the transformer on all metrics (Table~\ref{table:oad-scenario1-results}). In the transformer and TempcNN confusion matrices (Fig. \ref{fig:oad-results-transformer2}, \ref{fig:oad-results-tempcnn2}) high misclassification rates among the classes is apparent. In this scenario the LSTM architecture presented the same pattern with scenario 1, while the transformer performance was severely degraded. The adaptation of LSTM can be attributed to the fact that time is encoded as part of the feedback process, while transformers handle data as a sequence. This subtle difference on the time representation can have significant impact on the classification performance (transferring from one region to another), because the phenological cycles do not simply shift across time, but are also stretched or shrunk down depending on the soil-climate conditions of the area.

\subsubsection{Scenario 3}

In the third scenario, training was performed on France 2019 data and tested on Catalonia 2020 data. Similarly to the previous scenario, the LSTM model performed better than the transformer and TempCNN models, though not with the same performance gap (Table~\ref{table:oad-scenario1-results}). This degradation in performance can be attributed to the fact that in this scenario the training samples are drawn from a different distribution with respect to the test samples. The variation includes two components: the first component is that the growing season across the regions is different for the same crops (mainly due to different climatological conditions). The second component is that different years have slightly different starting dates for crop cultivation. Jointly these factors may significantly affect the classification performance.
    \section{Discussion}

Agricultural monitoring mainly includes tasks like crop type classification, parcel extraction and counting, and crop phenology evolution. Consistent acquisition of satellite multispectral time series of images are the stepping stone towards a complete agricultural monitoring process. The opening to the public of various LPIS systems, which contain crop type declarations and parcel geometries, bundled with the availability of consistent satellite measurements and advances in DL architectures enabled the creation of Sen4AgriNet. The first part of the challenge in order to materialize the dataset was the harmonization of the contextual information contained in the different LPIS. This challenge was solved by adapting a crop type classification structure inspired by FAO. The second part of the challenge was to build the dataset for multiple countries, bundling thousands of images with the corresponding labels in order to create a Machine Learning ready dataset. The challenge was met by creating a custom pipeline that automatically identifies the Sentinel-2 images that need to be downloaded from ESA’s Scihub \cite{esa_api_hub} and/or AWS Sentinel-2 bucket \cite{aws_s2_bucket}, downloads them, splits them to non-overlapping patches, stacks the different time stamps and finally attaches the labels. This procedure runs in an end to end fashion until the generation of the complete standalone netCDF4 files. 

\begin{figure*}[ht]
    \centering
    \subfloat[]{\includegraphics[scale=0.25] {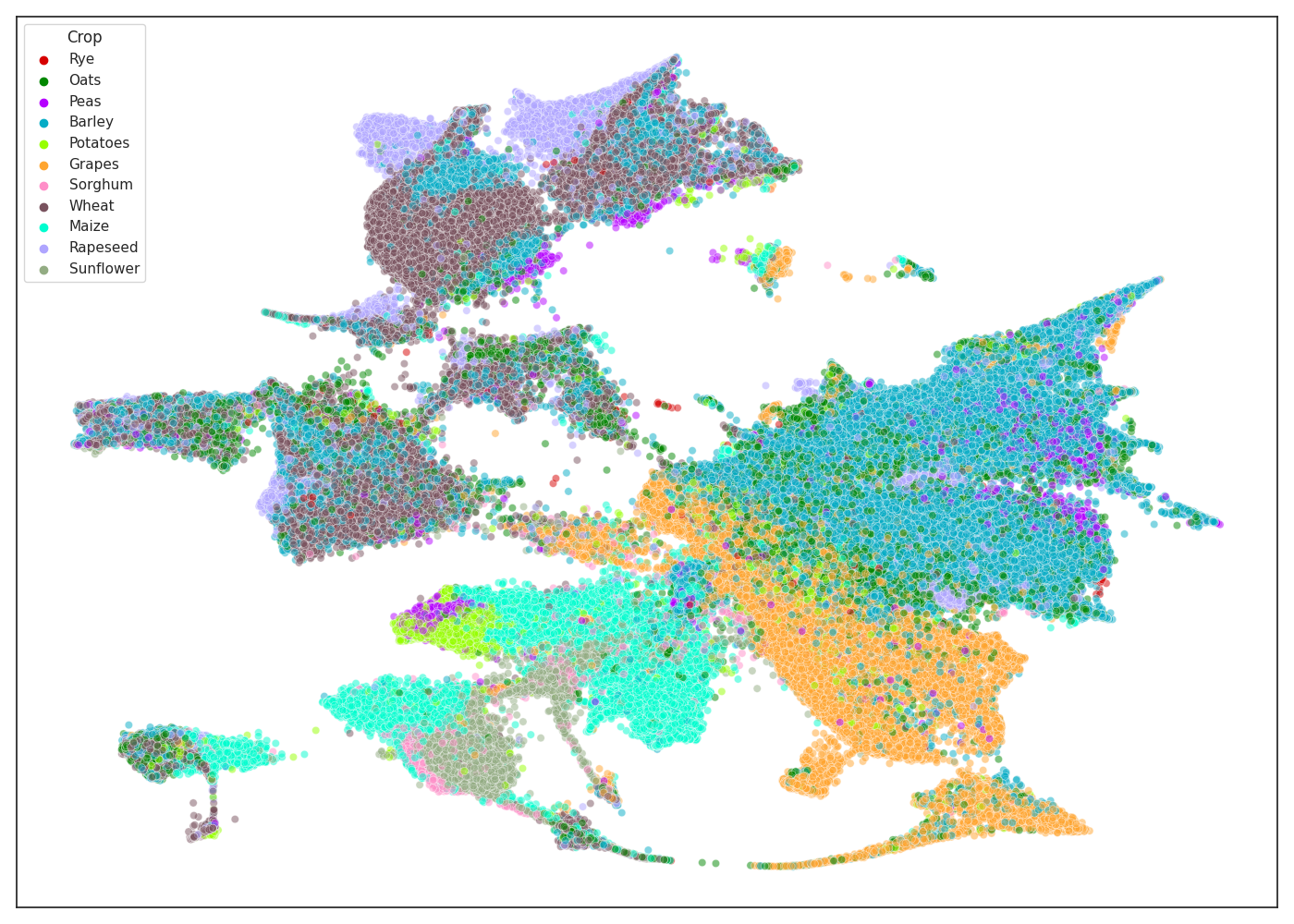}
    \label{dataset:umap_single}}
    \hfil
    \subfloat[]{\includegraphics[scale=0.25] {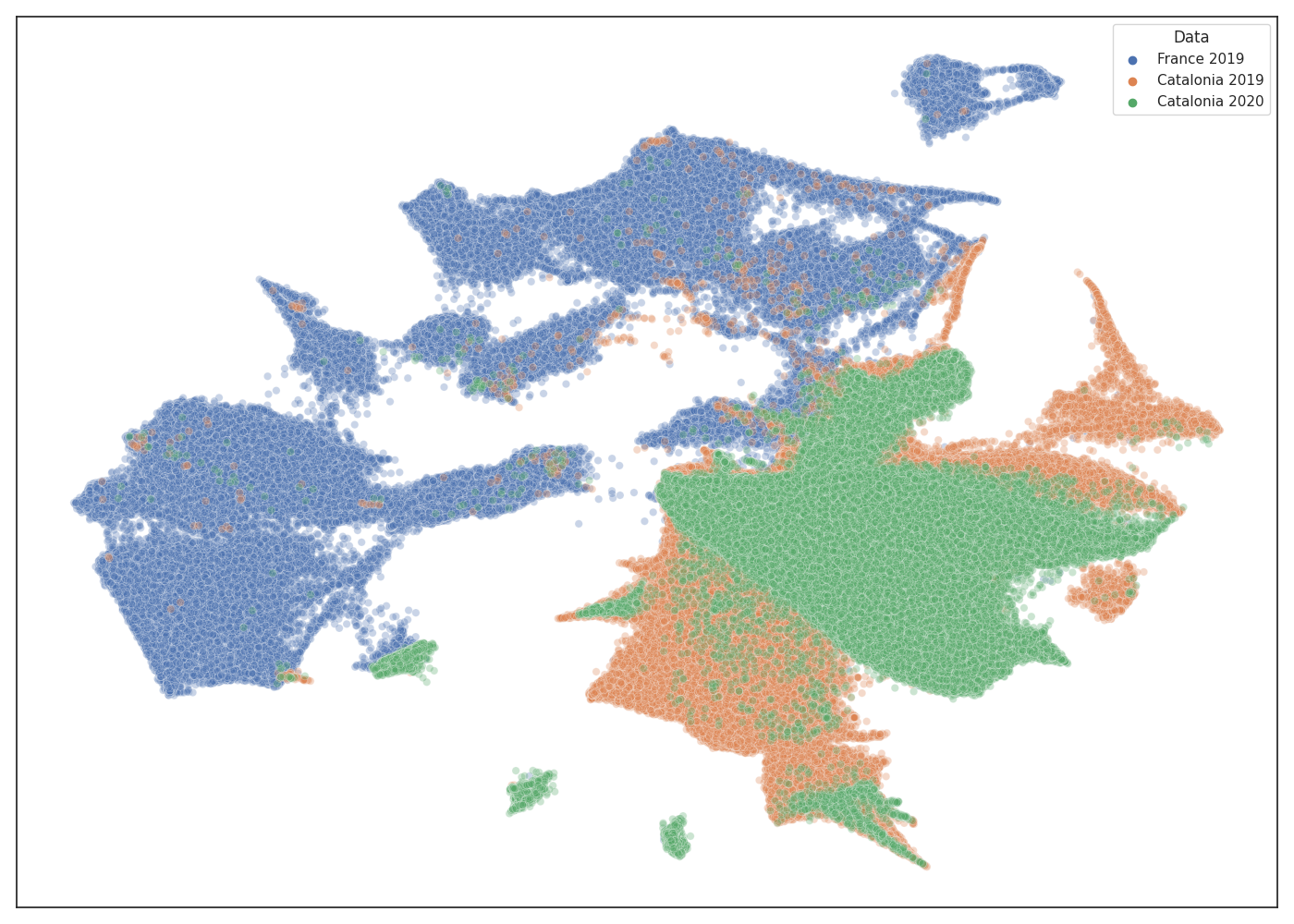}
    \label{dataset:umap_all}}
    \hfil
    \subfloat[]{\includegraphics[scale=0.16] {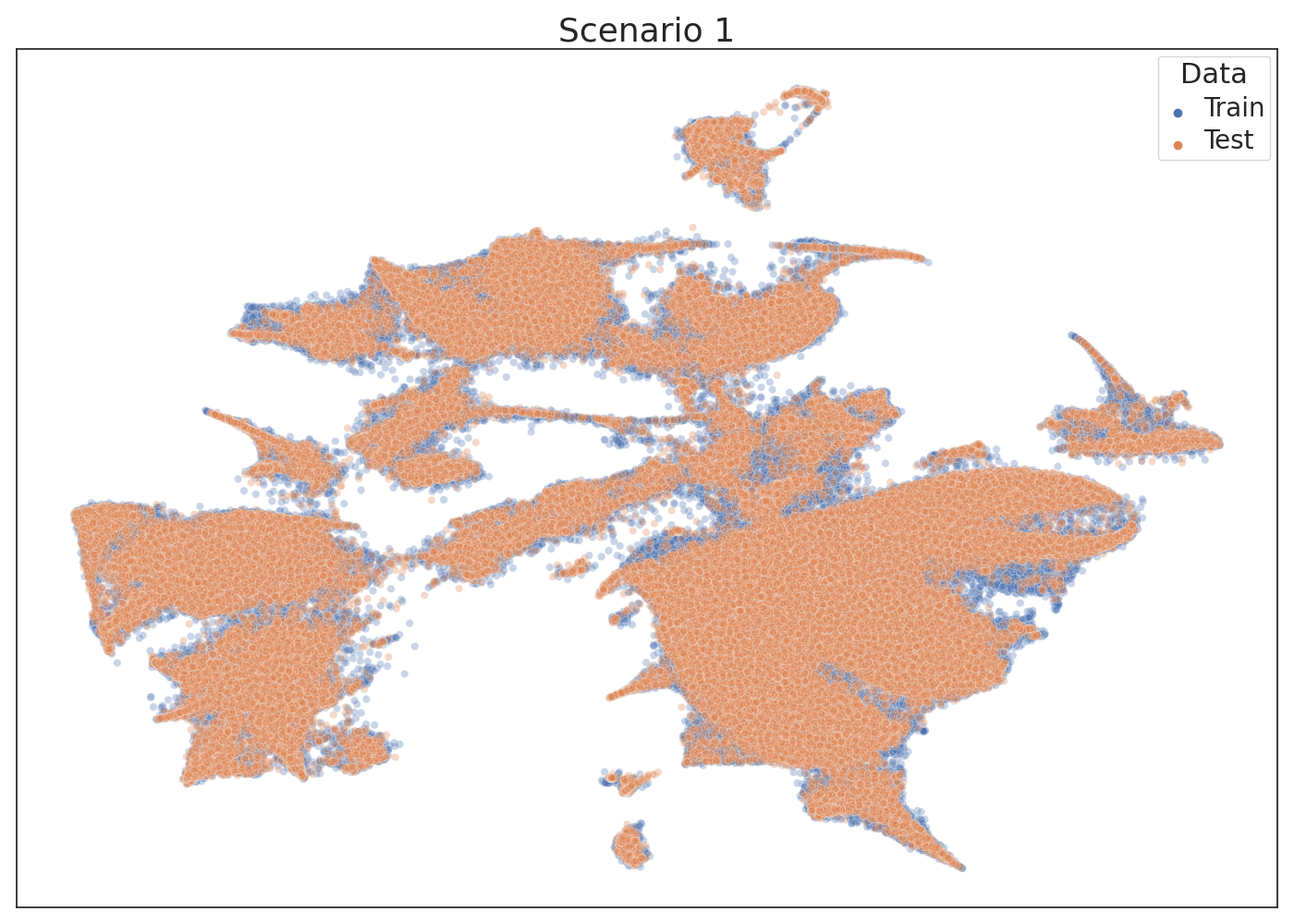}
    \label{dataset:umap_all}}
    \hfil
    \subfloat[]{\includegraphics[scale=0.16] {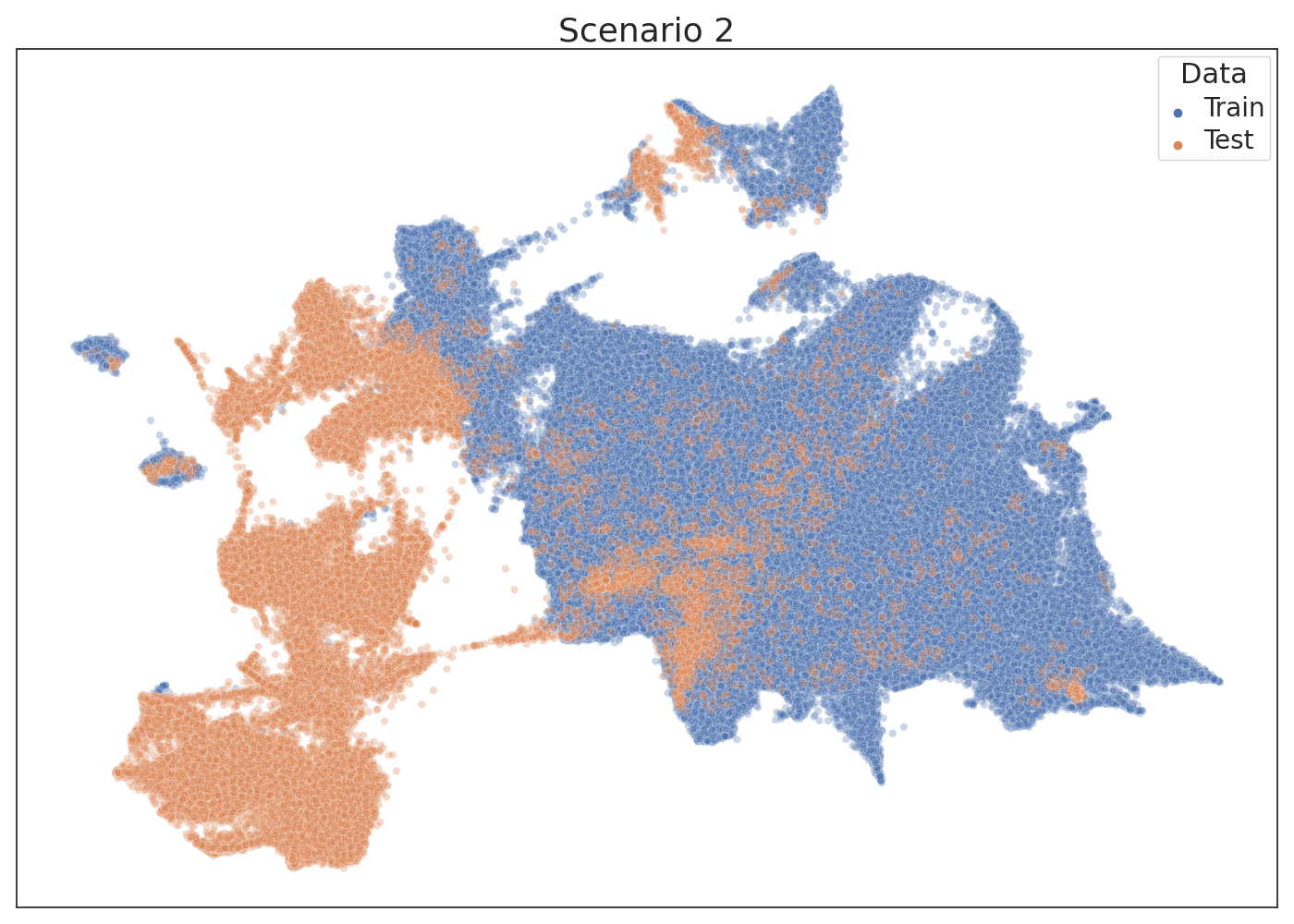}
    \label{dataset:umap_all}}
    \hfil
    \subfloat[]{\includegraphics[scale=0.16] {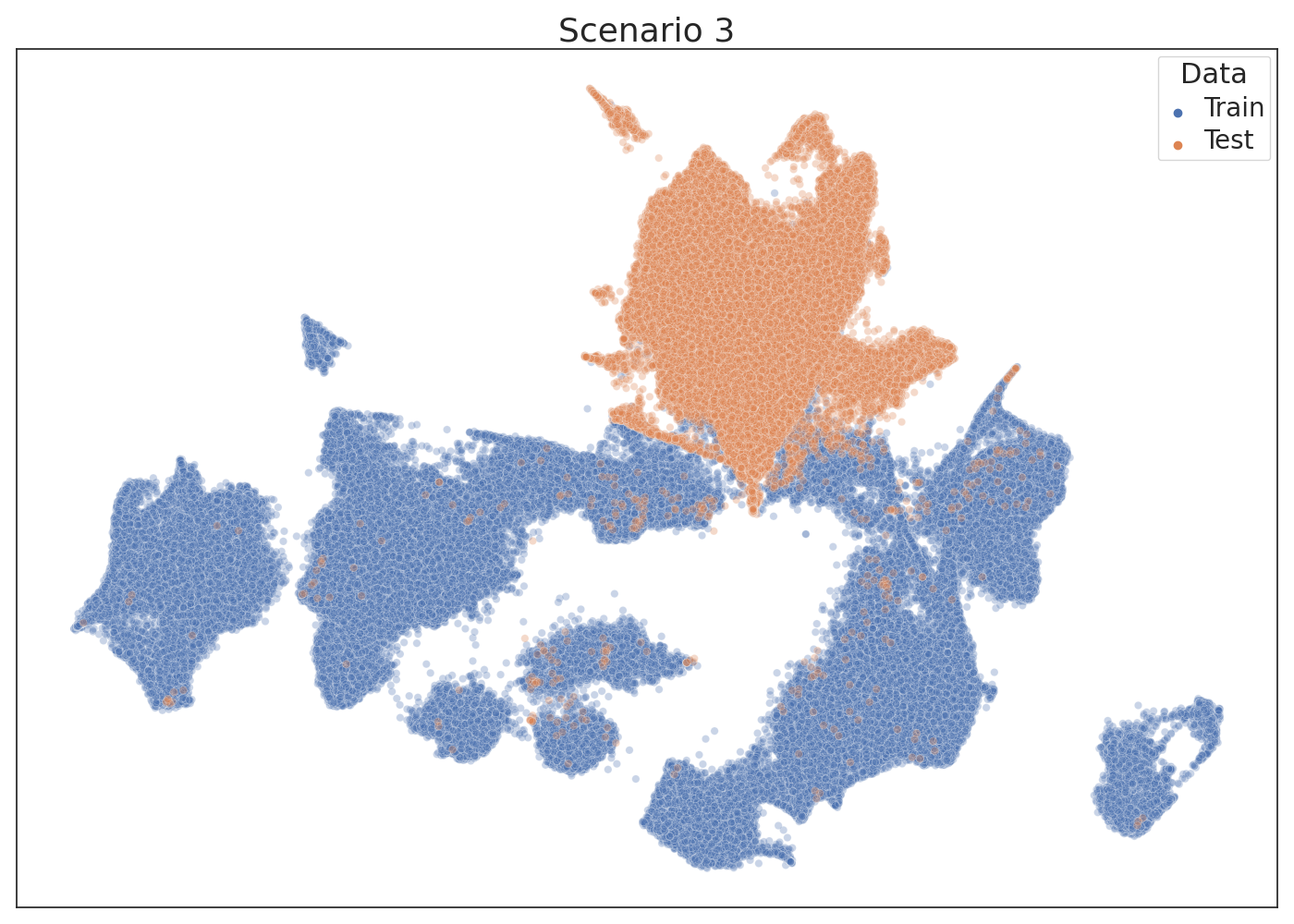}
    \label{dataset:umap_all}}
    
    \caption{UMAP visualizations for the examined datasets. (a) Distribution of the selected crops in all regions and years, (b) distribution of data across the three different datasets, (c)-(d) distribution of the train and test data in the three scenarios. Better viewed in colour.}
    \label{dataset:umap}
\end{figure*}

\begin{figure*}[htb]
    \centering
    \includegraphics[width=1.0 \textwidth]{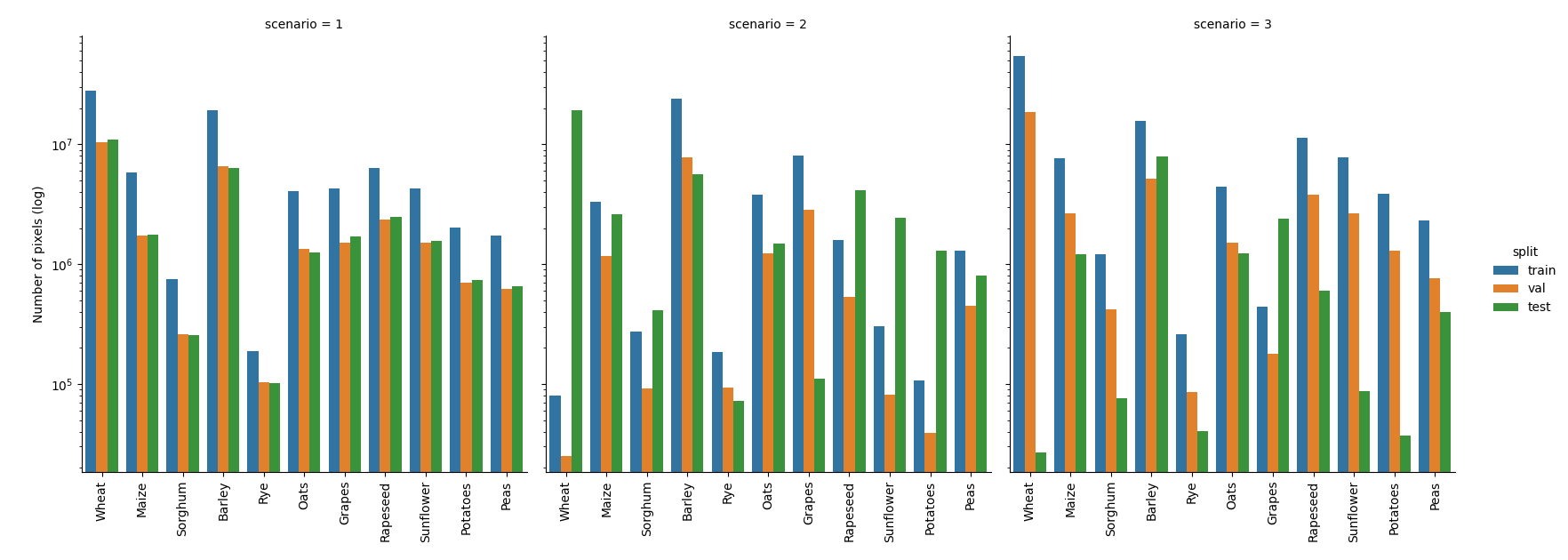}
    \caption{The pixel distribution of the selected crop classes across the different scenarios. Vertical axis is in logarithmic scale.}
    \label{fig:distribution}
\end{figure*}

A subsequent step involves the creation of appropriate splits for training, validation and test, which is a non-trivial task. We implemented a stratified sampling procedure, based on the location, the class and cultivation year of the parcels in our dataset, selecting those that have a significant number of observations and are common in both countries. However, this balanced selection of parcels in our splits, which is different for each experimental scenario, is not necessarily mapped to a balanced selection of pixels instead of parcels. In Fig.~\ref{fig:distribution} we show the pixel distribution of the different crop classes, for the three splits and for the three scenarios. Interestingly, in scenario 1 the balance parcel distribution is also mapped in the pixels distribution. In scenario 2 and 3 though, this is not the case. For example, in scenario 2 the number of wheat pixels are two orders of magnitude less in the train set than in the test set. The inverse is observed in scenario 3. These discrepancies are due to the different agricultural practices between Catalonia and France. In Catalonia agriculture is more fragmented and the average parcel area is less in hectares compared to France. Therefore balancing the splits based on parcel number is a good selection for the OAD set, while balancing the splits based on pixel count is a good selection for the PAD set.

Investigating the generalisation potential of DL models vis-à-vis the three different scenarios and the two different datasets, there are a couple of observations. First, in scenario 1, the probability density functions of the train and the test set should be nearly identical, since in both sets the samples experience the same spectral, spatial, geographic and temporal variability. This does not apply for scenario 2 and 3. In scenario 2 the temporal variability is captured but not the geographic/climatic. In scenario 3 both the yearly and the climatic variability information is lost. Therefore the probability density functions of the train and the test sets differ significantly. This explains the observed degradation in performance for both PAD (Table~\ref{table:pad-scenario1-results}) and OAD (Table~\ref{table:oad-scenario1-results}) datasets. In order to fortify this observation, different UMAP \cite{mcinnes2018umap} visualizations are provided in Fig. \ref{dataset:umap} which were computed using the OAD data. In Fig. \ref{dataset:umap}(a) a sample of the selected crops from all regions and years is visualized and we can see that most crops form distinct clusters, with some observed and anticipated confusion between those with similar phenology, e.g. wheat and rapeseed, oats and barley, etc. Fig. \ref{dataset:umap}(b) provides a visualization of the three different scenarios, where we observe a partial overlap between data from both years in Catalonia and a clearer divergence between Catalonia and France clusters. This indicates that both the temporal and more importantly the spatial diversity in Sen4AgriNet result in different data distributions. Finally, in Fig. \ref{dataset:umap}(c)-(e) the deviations between the train and test sets of the different scenarios are illustrated. As expected, the two splits in scenario 1 are heavily overlapping, in contrast to those in scenarios 2 and 3 where separate clusters are formed with some minor overlap, due to the different year and/or location used in the train and test splits.

Second, the OAD based experiments provide better results with respect to PAD, for all scenarios, especially for the second and third scenario. The enhanced classification performance and better generalisation of the OAD was expected. OAD aggregates several pixels for each parcel and the mean and standard deviation of this aggregation are included as input features. On the contrary, the PAD architectures used pixel based processing, thus the spatial heterogeneity is not well captured. In principle, our object based analysis smooths out the probability distributions in the spatial and temporal domain, removing high-frequency components. In the extreme case of training on 2019 France data and testing on 2020 Catalonia data (scenario 3), this translates to bringing the two data distributions closer. However, PAD and OAD solve different DL problems and direct comparison is not entirely fair. OAD's increased classification accuracy comes at the expense of spatial resolution, while it completely misses out on capturing inter-parcel variability. Finally, it is understood that the main obstacle in further improving the generalization capacity of a trained model is the inclusion of additional regions and with higher yearly coverage, to capture the characteristics of non dominant classes with more observations and extract more meaningful features. 

Regarding the PAD experiments, top accuracy was achieved as anticipated in the first scenario where regions and years are mixed, allowing the models to learn all temporal crop variations for both Catalonia and France. The most challenging classes to distinguish in scenario 2 are cereal-based crops, such as wheat, maize, sorghum, barley. This is attributed to their similar spectral content in the satellite time series. Finally, the third scenario experienced the worst performances, especially for ConvSTAR, which seems to be unable to discern the different classes with the sole exception of maize and rapeseed. Overall, U-Net shows more stable performance and manages to outperform the other two models in the most difficult scenarios (2 and 3) while achieving competitive results in the first scenario. This implies that the recursive nature of the ConvLSTM and ConvSTAN models does not offer much improvement in the studied cases. In addition, although ConvSTAR has considerably fewer training parameters and converges faster, this seems to inevitably come at the expense of accuracy. Lastly, ConvLSTM and U-Net employ a more complex encoder-decoder architecture, whereas ConvSTAR is a simple structure of three stacked ConvSTAR cells. According to the results of the different scenarios, this encoder-decoder scheme seems to be more efficient and achieve more robust results. ConvLSTM and U-Net manage to predict more compact crop areas with less internal variability, whereas ConvSTAR is more prone to predict multiple classes in a single parcel (Figs.~\ref{scenario1:vis_convlstm_pad},~\ref{scenario2:vis_convlstm_pad}, and~\ref{scenario3:vis_convlstm_pad}). The experiments suggest that ConvSTAR could benefit from an encoder-decoder architecture similar to ConvLSTM or a more thorough hyperparameter tuning. More complex models like  DuPLO \cite{INTERDONATO201991}, BCDU-Net \cite{Azad_2019_ICCV} or TempCNN \cite{rs11050523} may be more robust to geographical/temporal differences and are left as a future plan for experimentation. As a last thought, further exploration and finetuning of the models' hyperparameters could potentially boost the accuracy of the predictions and help produce higher quality classification maps.

On the OAD experiments, all model architectures have unique structures. This results in different training challenges; RNN networks and transformers are notorious for their high training time requirements. The training time required for each epoch is quite lower for the transformer-encoder classifier due to i) the tremendous difference in the number of total trainable parameters and ii) the limited parallelism in the nature of LSTM networks. However, despite the lower training time of transformer, in all training runs the LSTM model seemed to converge a lot faster requiring fewer epochs. Training time is not a issue when dealing with CNN based models, as parallelism is quite ubiquitous. This enables for far better training times and thus quicker convergence of the overall architecture.

Concerning the OAD reported results, the tests indicate that LSTM seems to perform better than the transformer. The transformer-encoder classifier seems to match the results of the LSTM in the first scenario, but it degrades dramatically in scenario 2. Transformers are considered as especially ``data hungry'' models, therefore using more data from Sen4AgriNet core dataset would provide a boost in the metrics. 
Finally, scenario 3 seems to be the middle ground for OAD, which was unexpected. This may be attributed to the ``bidirectional'' element of the LSTM cells which results in better performance since it enables them to learn in both directions. 

The above analysis for the experiments conducted in this study suggests that direct application of a trained DL model in a different setting hinders performance and greatly affects the generalizability of the model, thus some adjustment is required beforehand. The discrepancies in the label distributions, along with the intrinsic spatiotemporal variation observed between train and test sets in the different scenarios, provide an ideal setting for the exploration of Domain Adaptation techniques. Especially for scenarios 2 and 3, the direct knowledge transfer between source and target domains results in a significantly degraded performance and low quality predictions. In addition, the interested researcher can further design new scenarios with different types of spatiotemporal variation, e.g. by varying only the temporal dimension for the same region or by employing different sets of labels for train and test. In any case, careful adaptation of the model from one domain to another will certainly bridge this data distribution gap and improve performance and generalizability.

Last but not least, the extension of Sen4AgriNet to other types of segmentation tasks is straightforward. Apart from the semantic segmentation task presented in this study, instance segmentation and panoptic segmentation problems can also be addressed. In our case, instance segmentation refers to the localization of a parcel with a surrounding box and the pixels belonging to this parcel and specific crop type, managing to differentiate between adjacent parcels with the same crop. Panoptic segmentation extends the tasks of instance and semantic segmentation by including labelling of the surrounding environment, e.g. forests, water bodies, wetlands, etc. Sen4AgriNet comes with a total of 168 labels including non-crop related classes which can serve as the broad ``stuff''  category in panoptic segmentation terminology, whereas parcel boundaries can assist the identification of each individual object in an image for both tasks. We believe that the proposed dataset will encourage further research in this area and provide the required momentum for the development of more models and approaches, especially as far as panoptic segmentation is concerned where research studies focused on Remote Sensing are few and far between (e.g. \cite{de_carvalho_panoptic_2021}, \cite{garnot_panoptic_2021}).

    \section{Conclusion}

The lack of harmonized labeled data among the Paying Agencies, which operationally gather country wide labels every year, initiated the creation of Sen4AgriNet. Inspired by the FAO ICC nomenclature and adapted from the CAP and the remote sensing perspective, we proposed in this work the Sen4AgriNet crop type classification scheme. We build the benchmark dataset by leveraging the availability of Sentinel-2 multi-temporal, multi-country labeled data exploiting the recent opening up of LPIS parcel data. We construct a dataset that consists of 225,000 patches with corresponding pixel-based crop type maps. Based on this, we extracted two Sen4AgriNet subsets (PAD and OAD) for tackling different sets of classification problems, for unknown and known parcel geometries respectively. The experiments were divided into three different scenarios to investigate the impact of diverse agricultural practices, climatic zones, phenology phases, crop spectral signatures across different regions and cultivation years. As expected, the changes of the probability distribution functions experienced when moving between geographic regions for training and testing DL models, has a significant impact on classification performance and limits the models' capacity to adapt and generalize. 

We believe that Sen4AgriNet can be regarded as a labeled benchmark dataset, tailored for CAP and the use of Sentinel-2 imagery that come at no cost, and can spur numerous DL-based applications for crop type classification, parcel extraction, parcel counting and semantic segmentation. More importantly, the dataset can be extended to include other input data sources, including Sentinel-1 Synthetic Aperture Radar data, and meteorological data, allowing a new family of applications on early warning risk assessment and agricultural insurance.

    \section*{Acknowledgments}
    This work has received funding from the European Union’s
Horizon2020 research and innovation project DeepCube, under grant agreement number 101004188. 
    
    \bibliographystyle{IEEEtran}
    \bibliography{references}
    
    \vfill

\end{document}